\documentclass[twocolumn]{svjour3}
\smartqed  
\usepackage{graphicx}
\usepackage{amssymb}
\usepackage{float}
\usepackage{amsmath}
\usepackage{algorithm2e,algorithmic}
\usepackage{subfigure}
\usepackage{url}
\usepackage{color}
\DeclareGraphicsExtensions{.png,.jpg}

\newcommand{\R}{I\!\! R}

\newcommand{\mf}{\mathbf}

\usepackage{dsfont}
\def\one#1{\mathds{1}_{#1}}

\begin{document}

\title{A linear framework for region-based image segmentation and inpainting involving curvature penalization}
\author{Thomas Schoenemann$^1$, Fredrik Kahl$^1$, Simon Masnou$^2$ and Daniel Cremers$^3$}

\institute{$^1$Center for Mathematical Sciences, Lund University, Sweden.\\
          $^2$Universit\'e de Lyon, Universit\'e Lyon 1, CNRS UMR 5208, Institut Camille Jordan, 43 boulevard du 11 novembre 1918, 69622 Villeurbanne Cedex, France.\\
                $^3$Department of Computer Science, TU M\"unchen, Germany.}

\maketitle

\begin{abstract}
We present the first method to handle curvature regularity in
region-based image segmentation and inpainting that is independent of
initialization.

To this end we start from a new formulation of length-based
optimization schemes, based on surface continuation constraints, and
discuss the connections to existing schemes. The formulation is based
on a \emph{cell complex} and considers basic regions and boundary
elements. The corresponding optimization problem is cast as an integer
linear program.

We then show how the method can be extended to include curvature
regularity, again cast as an integer linear program. Here, we are
considering pairs of boundary elements to reflect curvature. Moreover,
a constraint set is derived to ensure that the boundary variables
indeed reflect the boundary of the regions described by the region
variables.

We show that by solving the linear programming relaxation one gets
quite close to the global optimum, and that curvature regularity is
indeed much better suited in the presence of long and thin objects
compared to standard length regularity.
\end{abstract}

\section{Introduction}

Regularization is of central importance for many inverse problems in
computer vision including image segmentation and inpainting
\cite{Mumford-Shah-89,Chan-Vese-01,Masnou-Morel-98,Bertalmio-et-al-00,Tschumperle-06,Bornemann-Maerz-07}.
The introduction of higher-order regularizers in respective energy
minimization approaches is known to give rise to substantial
computational challenges.  Some of the most powerful approaches to
image segmentation are based on region integrals with regularity terms
defined on the region boundaries
\cite{Blake-Zissermann-Book-87,Mumford-Shah-89,Nitzberg-Mumford-Shiota-93,Boykov-Jolly-01,Chan-Vese-01,Esedoglu-March-03,Klodt-et-al-08}.
While many such methods make use of length as a regularity term, only
few use curvature regularity. This is in contrast to psychophysical
experiments on contour completion \cite{Kanizsa-71} where curvature
was identified as a vital part of human perception. Note that
curvature regularity is qualitatively different from length
regularity. As shorter boundary curves are preferred in the length
case, this causes the well-known shrinking bias. This is not the case
for curvature, since the total curvature of any closed, convex curve
is equal to $2\pi$.

\begin{figure*}
\begin{center}
 \setlength{\tabcolsep}{1mm}
\begin{tabular}{cccc}
 \includegraphics[width=0.235\textwidth]{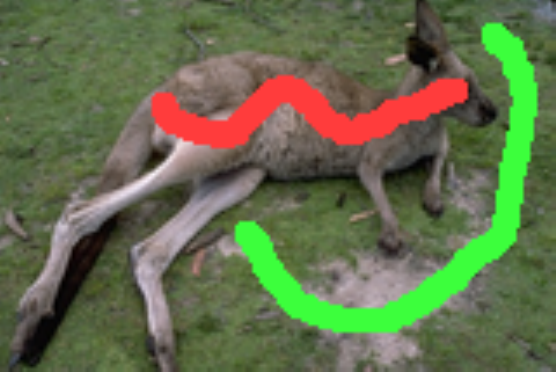} &
 \includegraphics[width=0.235\textwidth]{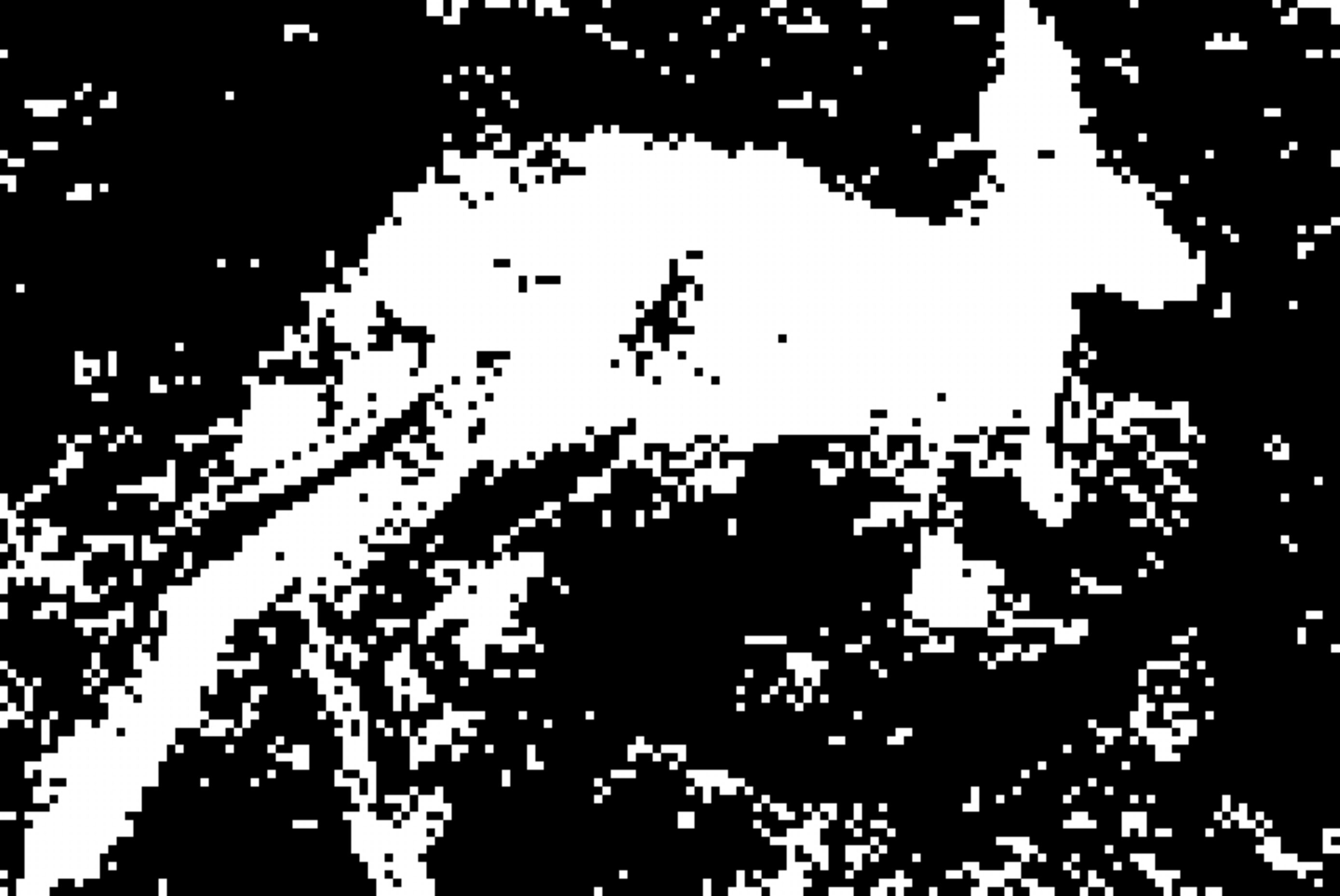} &
 \includegraphics[width=0.235\textwidth]{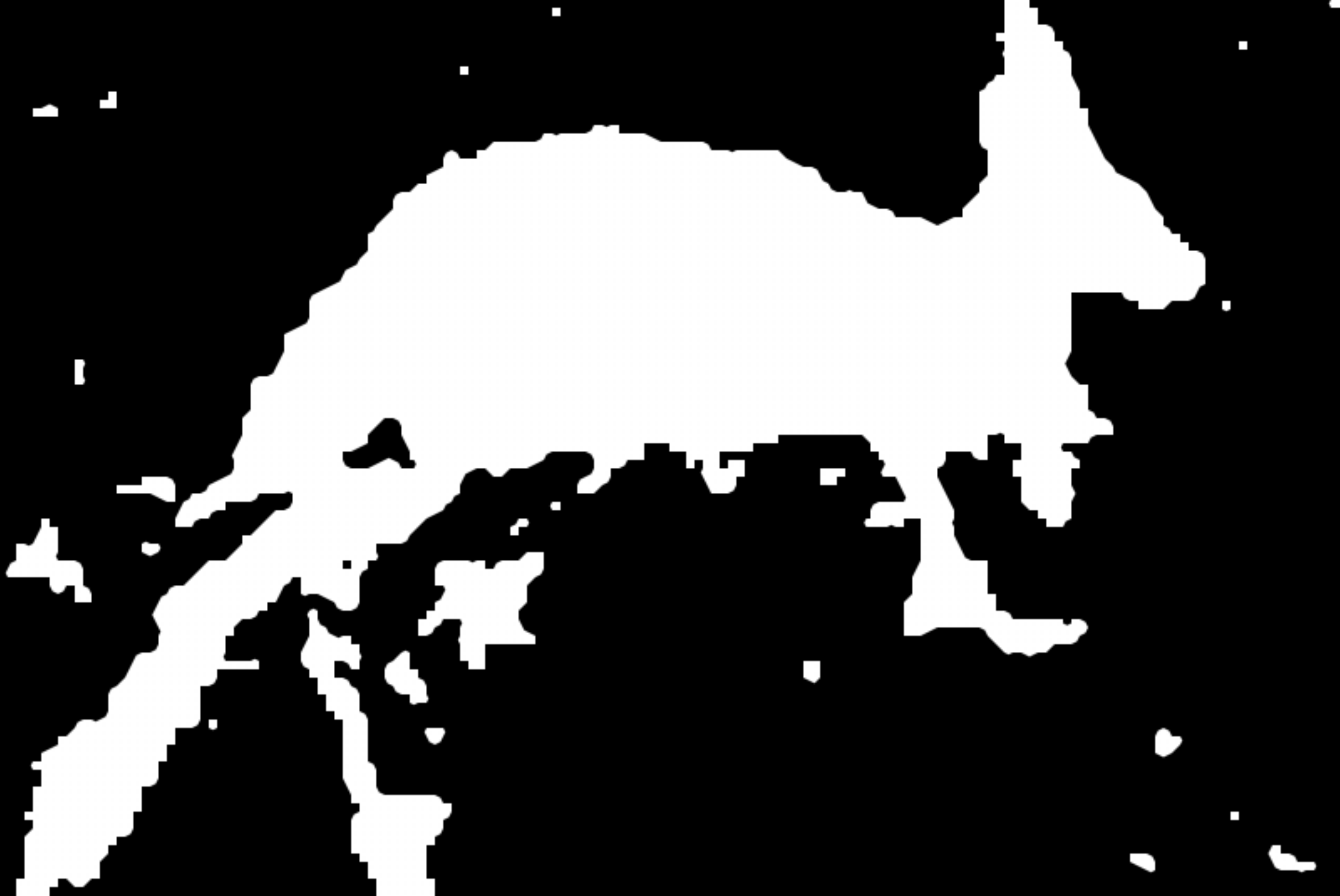} &
\includegraphics[width=0.235\textwidth]{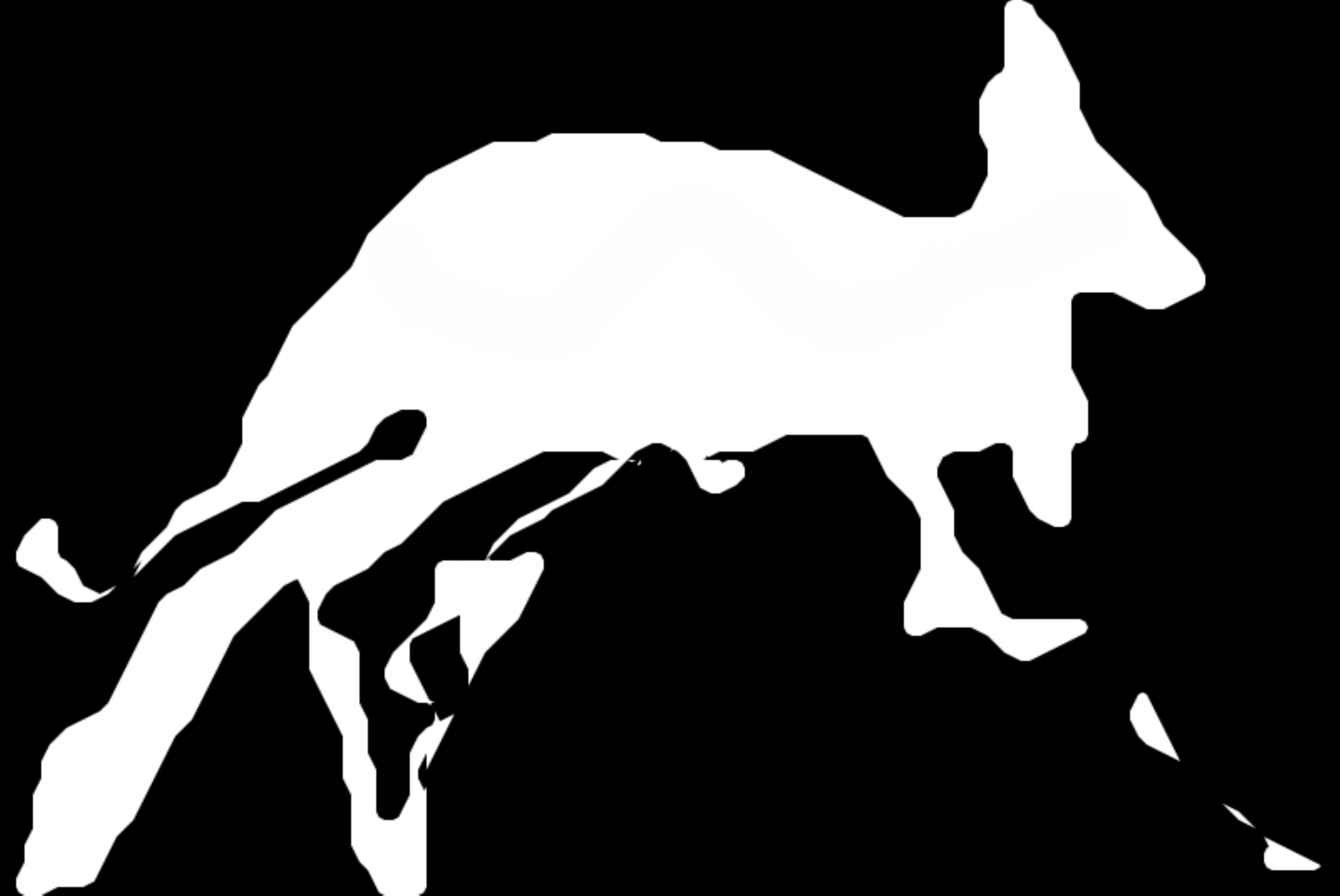}
 \\[-1mm]
 {\small Input with seeds} & {\small thresholding scheme} & {\small with length regularity}
& {\small with length and curvature}
\end{tabular}
\caption{With the proposed method, long and thin structures are much better handled than with length-based approaches.}
\label{fig:intro}
\end{center}
\end{figure*}

\begin{figure*}
\begin{center}  
\begin{tabular}{ccc}
\includegraphics[width=0.3\textwidth]{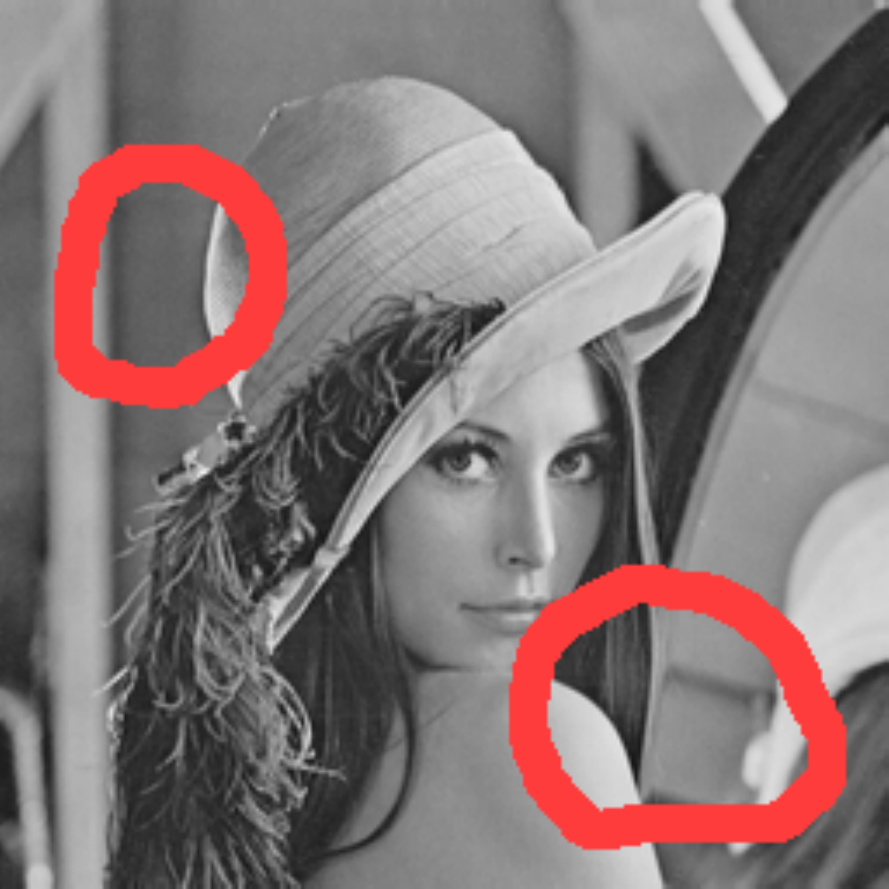}
& \includegraphics[width=0.3\textwidth]{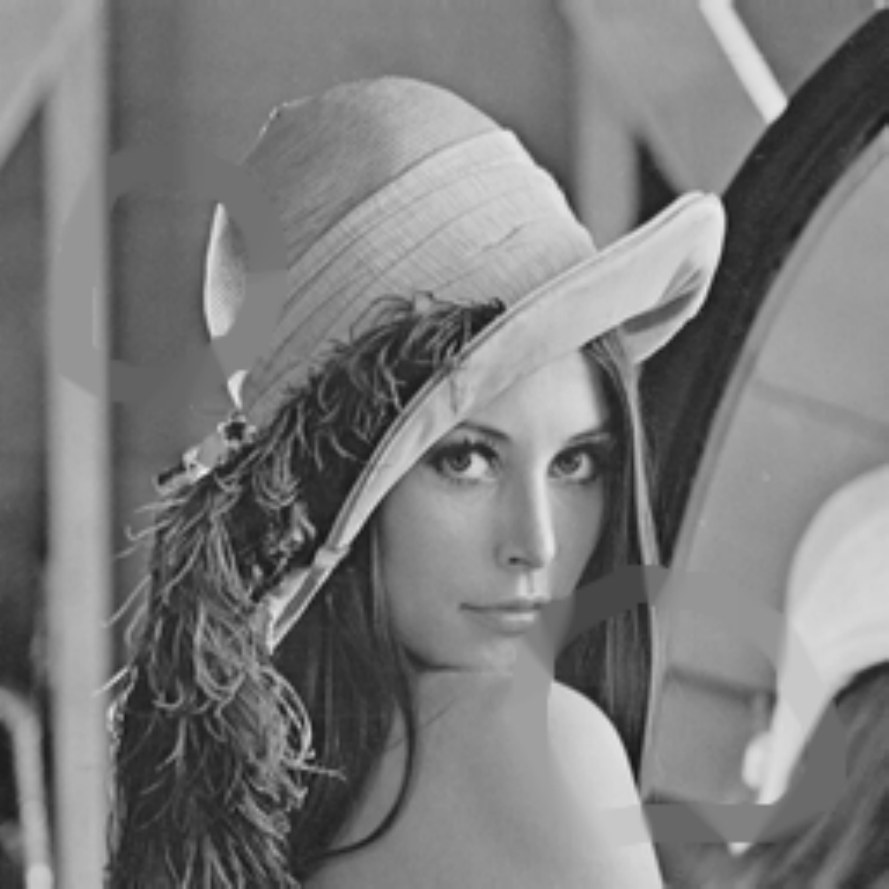}
& \includegraphics[width=0.3\textwidth]{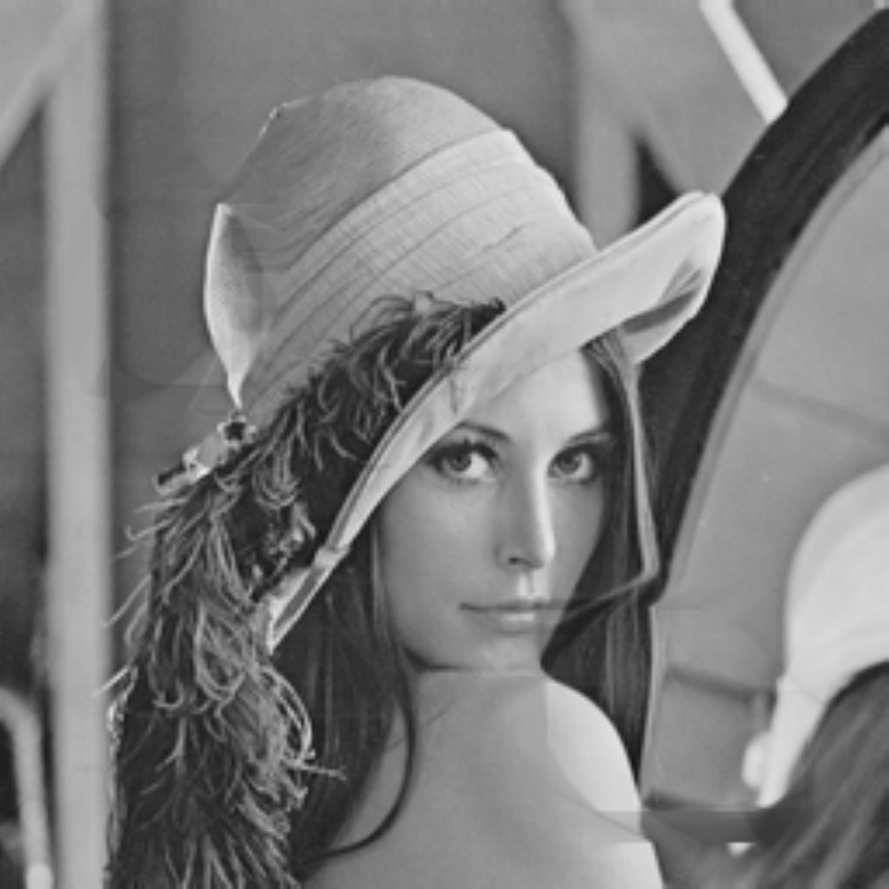}
\\
{\small damaged image} & {\small with length regularity}
& {\small with curvature regularity}
 \end{tabular}
 \end{center}
 \vspace{-1mm}
 {\bf
 \caption{{\em Curvature regularity} on the level lines improves inpainting.}
 \label{fig:intro2}
 }
\end{figure*}

Length regularization has become an established paradigm as there
exist many powerful algorithms for computing optimal solutions for
such energy functionals, either using discrete graph-theoretic
approaches based on the min-cut/max-flow duality
\cite{Greig-Porteous-Seheult-89,Boykov-Jolly-01} or using continuous PDE-based
approaches using convex relaxation and thresholding theorems
\cite{Nikolova-Esedoglu-Chan-06}.  Region-based problems for
segmentation using curvature regularity have typically been optimized
using local optimization methods only (cf.\
\cite{Nitzberg-Mumford-Shiota-93,Esedoglu-March-03}).  As a consequence,
experimental results highly depend on the choice of initialization.
Moreover, these methods do not offer any insights concerning
how close the computed solution is to the (unknown) global solution.

In this paper, we propose a relaxed version of region-based
segmentation which can be solved optimally.  The key idea is to cast
the problem of region-segmentation with curvature regularity as an
integer linear program (ILP). By solving its LP-relaxation and
thresholding the solution we obtain a solution to the original integer
problem and are able to evaluate a bound on its quality with respect
to the globally optimal solution.  In addition, we show that the
method readily extends to the problem of inpainting.

Figure \ref{fig:intro} demonstrates that the proposed method allows
segmenting objects in a way which preserves perceptually important
thin and elongated parts. The found solution is within $1.3\%$ of the
global optimum. Figure \ref{fig:intro2} demonstrates the superior
performance of curvature regularity over length regularity in a
corresponding inpainting experiment.

\paragraph{Existing Work on Curvature Regularity.}

For contour- or edge-based segmentation methods researchers have
successfully developed algorithms to optimally impose curvature
regularity using shortest path approaches \cite{Amini-et-al-90} or
ratio cycle formulations \cite{Schoenemann-Cremers-07b} on a graph
representing the product space of image pixels and tangent angles
\cite{Parent-Zucker-89}.  In the region-based settings considered,
curvature is usually handled by local evolution methods
\cite{Chan-Kang-Shen-02,Esedoglu-March-03,Nitzberg-Mumford-Shiota-93,Tschumperle-06}.
Among the methods that pre-date our conference publication
\cite{Schoenemann-Kahl-Cremers-09}, the only exception we are aware of
is the inpainting approach of Masnou and Morel \cite{Masnou-Morel-98}
who can optimize the $L_1$-norm of the curvature in the absence of
regional data terms using dynamic programming.

In this paper we propose an LP-relaxation approach to minimize
curvature in region-based settings. In contrast to
\cite{Masnou-Morel-98} it allows imposing arbitrary functions of
curvature and arbitrary data terms. The algorithmic formulation is
based on the concepts of \emph{cell complexes} and
\emph{surface continuation constraints} which have  been pioneered by
Sullivan \cite{Sullivan-94} and Grady \cite{Grady-10} in the
context of 3D-surface completion.

The present paper is based on our preliminary work in
\cite{Schoenemann-Kahl-Cremers-09}, but contains several novelties.
Firstly, we show that the constraint system in
\cite{Schoenemann-Kahl-Cremers-09} needs to be augmented by additional 
constraints in order to ensure that the boundary of the region-based
segmentation is correctly estimated. Secondly, we discuss the
connections of our method, when restricted to length regularity, with
standard methods for length regularity and compare experimentally.  In
addition, the original inpainting method has improved by incorporating
a boundary estimation scheme.

There has been a subsequent paper by El-Zehiry and Grady
\cite{ElZehiry-Grady-10} which optimizes the same model as in our
original paper, but applies quadratic pseudo-Boolean optimization
(QPBO) for obtaining the solution. In the case of a square grid, QPBO
is able to efficiently compute a solution with curvature
regularization, and with a subsequent probing stage often the global
optimum is found. However, the discretization artefacts are severe for
a cell complex of squares. Better connectivities can only be handled
approximately.

The source code associated to this paper is freely available at 
\texttt{http://www.maths.lth.se/matematiklth/personal/\\tosch/download.html}.

\section{Length-based Segmentation Problems}

\label{sec:len}

To detail the proposed method for curvature regularity, we first
introduce a novel method for length-based segmentation problems.  In
practice there are more efficient algorithms for this problem
\cite{Boykov-Kolmogorov-03,Nikolova-Esedoglu-Chan-06}, but in
contrast to them the presented one is easily extended to curvature. A
comparison to the known techniques is given at the end of this
section.

Given an image $I:\Omega\rightarrow \R$, the problem is to segment it
into two regions, foreground and background. Here ``region'' means an
arbitrary subset of $\Omega$, i.e.\ there can be several disconnected
components and each one can have holes. Hence, each point $\mf{x} \in
\Omega$ is to be assigned a region $u(\mf{x}) \in \{0,1\}$ where $0$
denotes background, $1$ foreground.

The desired segmentation is defined as the global optimum of an energy
function, consisting of two terms. The first one is called the
\emph{data term} and specified by a function $g_0(\mf{x})$ for
points belonging to the background and a function $g_1(\mf{x})$ for
the foreground. Both functions will generally depend on the input
image $I$. In addition there is a \emph{regularity} term that
penalizes the length of the segmentation boundary by a weighting
parameter $\nu \ge 0$. The arising energy minimization problem to be
solved is
\begin{eqnarray}
&\min\limits_{u: \Omega \rightarrow \{0,1\}}& \int\limits_\Omega 
g_0(\mf{x}) [1 - u(\mf{x})]\, d\mf{x}\, + \, \int\limits_\Omega g_1(\mf{x})\, u(\mf{x})\, d\mf{x}
\nonumber\\
&& \hspace{1cm} +\, \nu\, |C|\ \ \ , \label{eq:continuous_length}
\end{eqnarray}
where $C = \partial \{\mf{x}\, |\, \nabla u(\mf{x}) = \mf{0} \}$ is
the set of points where $u$ is discontinuous (i.e.\ ``jumps'' from $0$
to $1$) and $|C|$ denotes its one-dimensional measure. In other words,
$C$ is a set of closed lines (which may include parts of the boundary
of $\Omega$) and $|C|$ denotes the sum of the length of all lines.

For convenience, we reformulate \eqref{eq:continuous_length} by
splitting the integrand of the first term into a constant term and one
depending on $u$. Defining $g(\mf{x}) = g_1(\mf{x}) - g_0(\mf{x})$ the
resulting functional is
\begin{eqnarray}
&\min\limits_{u: \Omega \rightarrow \{0,1\}}& 
\int\limits_\Omega g(\mf{x})\, u(\mf{x})\, d\mf{x}\ + \nu\, |C| 
\ +\ \mbox{const}\ \ .
\end{eqnarray}
In the following we will ignore the constant except when evaluating
relative gaps between a lower bound and the energy of some
segmentation.

\subsection{Discretization}

In this paper we consider \emph{discretized} segmentation problems
where instead of optimizing infinitely many values $\{u(\mf{x})\, |\,
\mf{x} \in \Omega\}$ we only consider finitely many ``basic regions''
and jointly assign all points in a basic region to the same
segment. Note that in practice we will always get a discrete input
image $I$, where the basic regions are given by pixels. Hence, the
data term by itself will produce such an assignment even for the
continuous problem. This is no longer true when the regularity term is
added, but in practice the discretized energy function can be designed
to account for this phenomenon
\cite{Boykov-Kolmogorov-03,Nikolova-Esedoglu-Chan-06}.

\begin{figure}
\begin{tabular}{cc}
\includegraphics[width=0.25\textwidth]{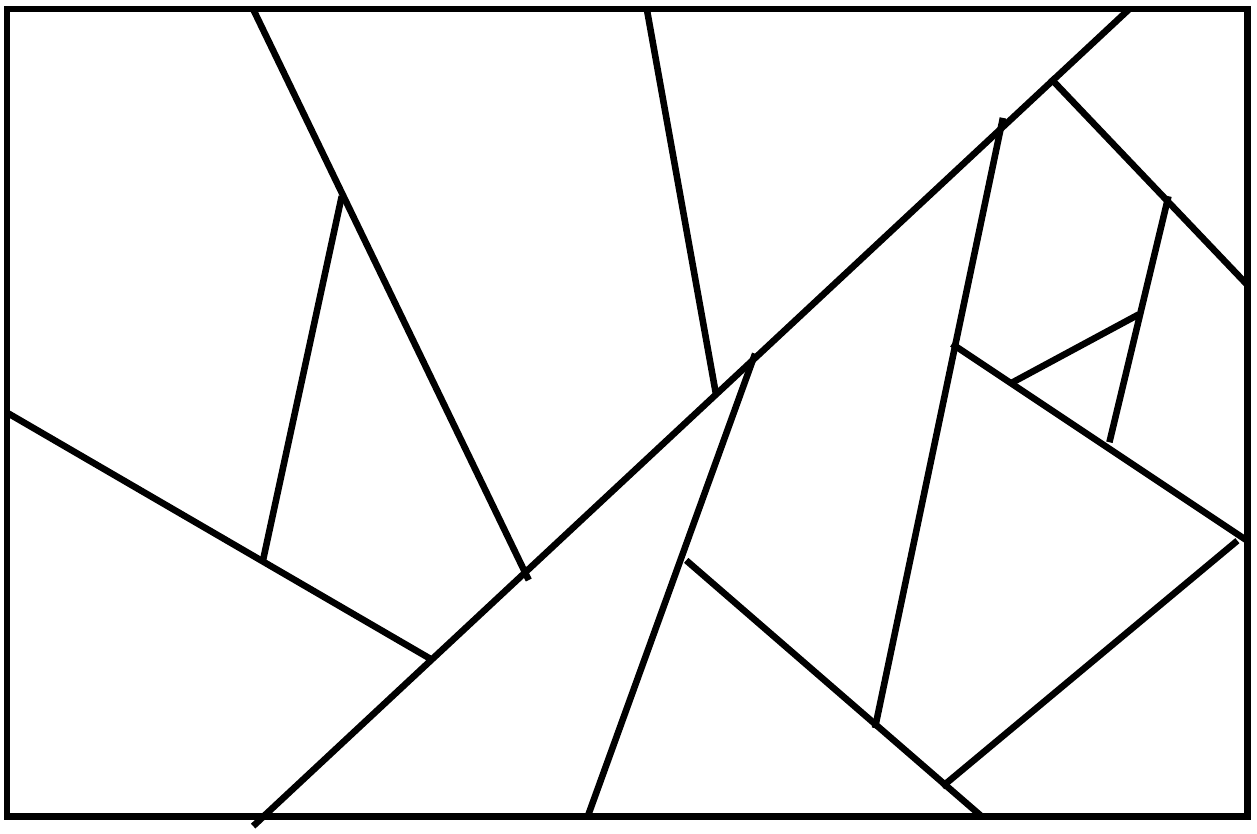} &
\includegraphics[width=0.175\textwidth]{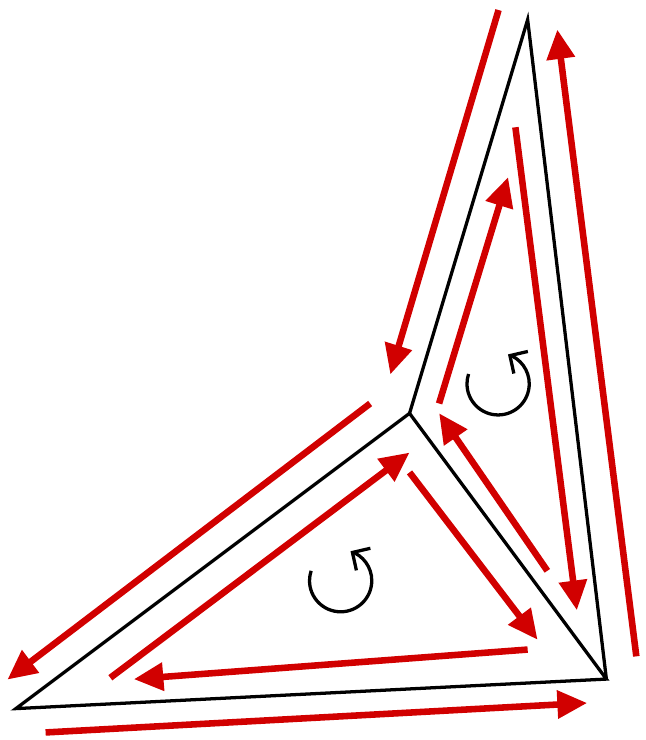} \\[-1mm]
{\small (a)} & {\small (b)}
\end{tabular}
\caption{The basic concepts of our method. (a) A cell complex. (b) 
The method considers oriented regions and oriented boundary elements.}
\label{fig:illustration}
\end{figure}

We require that our set of basic regions -- denoted $\mathcal{F}$ --
be a \emph{cell complex} and a partitioning of $\Omega$, i.e.\ that
(1) no two regions overlap and (2) the union of all basic regions
yields $\Omega$. An example is given in Figure \ref{fig:illustration}
(a).

The presented approach makes use of another essential part of a cell
complex: \emph{boundary segments}. These are the line segments that
form the borders of the basic regions. Usually a boundary segment has
two neighboring regions, except for segments at the border of $\Omega$
where there is only one. The set of all boundary segments is denoted
$\mathcal{E}$. As will be shown below we need to consider both
possible ways of traversing a boundary segment. Hence, for each
boundary segment we consider two \emph{oriented boundary segments},
also called \emph{line segments} in the following. The set of all
these segments is denoted $\mathcal{E}^O$ and $\ell(e)$ will denote
the length of a line segment $e$.

\begin{figure}
\begin{tabular}{cc}
\includegraphics[width=0.2\textwidth]{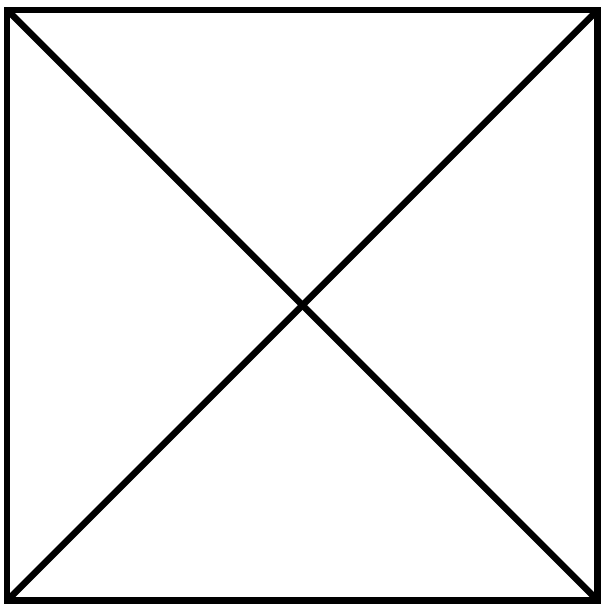} &
\includegraphics[width=0.2\textwidth]{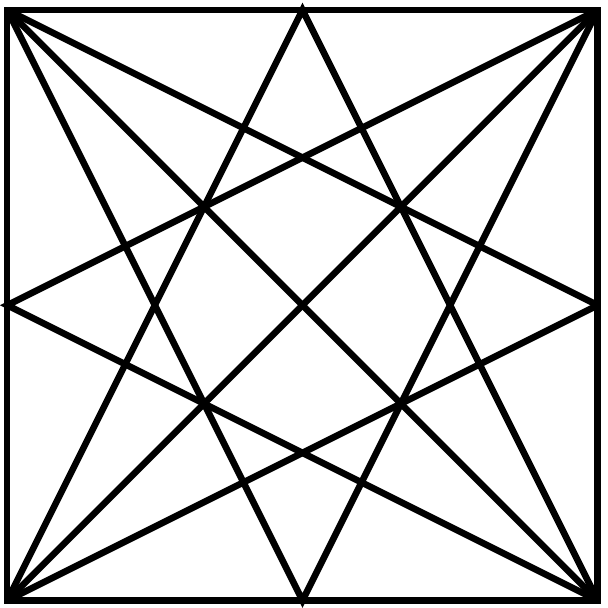}\\[-1mm]
{\small $8$-connectivity} & {\small $16$-connectivity}
\end{tabular}
\caption{Splitting a pixel into basic regions using lines with $8$ and $16$ different directions.}
\label{fig:pixel_splits}

\end{figure}

In essence, the data term will be defined in terms of the basic
regions, the regularity term in terms of boundary segments. To
approximate the continuous problem sufficiently well, basic regions
should generally not correspond to pixels. Instead, as detailed in
Figure \ref{fig:pixel_splits} we split the pixels into either $4$ or
$32$ basic regions, induced by lines with $8$ and $16$ different
directions respectively. In the following we will refer to this as
$8$- and $16$-connectivity.

\subsection{An Integer Linear Program}

The presented method casts a discretized version of
\eqref{eq:continuous_length} as a so-called \emph{integer linear
program}, i.e.\ minimizing a linear cost function over integral
variables and subject to linear constraints. There are two sets of
variables: firstly, for each basic region $f \in \mathcal{F}$ there is
a \emph{region variable} $\mf{y}_R^f \in \{0,1\}$ with $0$ indicating
that the region belongs to the background, $1$ to the foreground. The
second set contains a \emph{boundary variable} $\mf{y}_B^e\in \{0,1\}$
for every oriented boundary segment $e\in\mathcal{E}^O$. Here we want
$\mf{y}^B_e$ to be $1$ only if exactly one of the adjacent basic
regions belongs to the foreground.  Now we are already in a position
to express the cost function:
\begin{equation}
\phantom{xxxx} \mf{c}_R^T\, \mf{y}_R\ +\ \mf{c}_B^T\, \mf{y}_B \ \ ,
\label{eq:cost}
\end{equation}
where $\mf{c}_R$ contains entries
$$
\mf{c}_R^f = \int\limits_f g(\mf{x})\ d\mf{x}\ \ ,
$$ and $\mf{c}_B$ contains entries $c_B^e = \nu\, \ell(e)$. Note that
since in our case basic regions are always subsets of a single pixel
this weight is simply the function $g(\cdot)$ evaluated at the pixel
times the area of the basic region.

Since $\nu$ is positive minimizing \eqref{eq:cost} by itself would set
all boundary variables to $0$, so we need constraints. Indeed, up to a
few ambiguities (see next section) the region boundary is completely
specified by the region variables and the boundary variables serve to
render the cost function linear. They are forced to describe the
correct boundary by the linear constraint system that we now
describe. In words it can be stated as

\begin{quote}
\textbf{Surface Continuation Constraint:}
\textit{
Whenever a basic region is part of the foreground, along each of its
boundary segments the foreground must either continue with another
foreground region or with an appropriately oriented boundary segment.
}
\end{quote}

Formalizing these constraints (one for each boundary segment) involves
the concept of \emph{orientations} for both regions and boundary
segments.  For boundaries we have already introduced this concept, but
it is essential for the constraint system that we (arbitrarily) define
a ``positive'' and a ``negative'' orientation for each boundary
segment.

For a region, an orientation denotes one of the two possibilities for
traversing its boundary line - clockwise or counter-clockwise. Here it
is essential that all regions have the same orientation.

Now, to formalize the surface continuation constraint we define the
notion of \emph{positive} and \emph{negative} incidence of regions
$f\in \mathcal{F}$ and line segments $l\in \mathcal{E}^O$ to boundary
segments $e\in \mathcal{E}$. Ultimately the constraint will then state
that the weighted sum of ``active'' region incidences must be equal to
the weighted sum of ``active'' boundary incidences, where
\emph{active} refers to elements where the associated indicator
variable is $1$.

The incidence of a region $f$ to a boundary segment $e$ is denoted
$m_e^f$ (for ``match'') and defined as $0$ unless $f$ contains $e$ in
its boundary line. Otherwise, $m_e^f$ is $1$ if according to the
orientation of $f$ the segment $e$ is traversed in its positive
orientation, otherwise $-1$ for the negative orientation. The
incidence of a line $l$ and a boundary segment $e$ is denoted $m_e^l$
and defined as $0$ unless $l$ is an orientation of $e$, otherwise $1$
if $l$ is the positive and $-1$ if $l$ is the negative orientation of
$e$. The constraint system can now be stated as
\begin{equation}
\phantom{xxxx} \sum\limits_{f\in \mathcal{F}} m_e^f\, y_R^f\ =\ 
\sum\limits_{l\in \mathcal{E}^O} m_e^l\, y_B^l\ \ \ \forall e \in \mathcal{E}\ \ \ .
\end{equation}
For clarity, let us look at the following example:

\vspace{5mm}
\includegraphics[width=.4\textwidth]{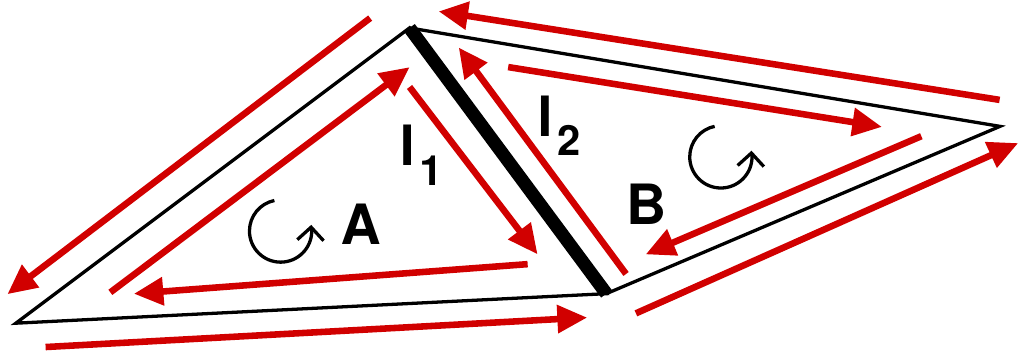}
\vspace{5mm}

\noindent Here the constraint for the bold edge reads 
\begin{equation}
\phantom{xxxx} y_R^A - y_R^B\ =\ y_B^{l_2} - y_B^{l_1}\ \ .
\label{eq:len_surf_cont}
\end{equation}
In the case where $y_R^A = 1$ and $y_R^B = 0$, i.e.\ region $A$ is
foreground and region $B$ is background, $y_B^{l_2}$ will be forced to
$1$, whereas $y_B^{l_1}$ will be set to $0$.  If instead $B$ is
foreground and $A$ background, this will force $y_B^{l_1}$ to be $1$
and $y_B^{l_2}$ to be $0$. If both $A$ and $B$ belong to the same
component, the constraint leaves some freedom for the boundary
variables: they can now both be $0$ or both be $1$. The latter is
undesirable, but will not happen as long as the length weight is
strictly positive. However, when we integrate curvature below we will
need extra constraints to prevent this case.

Finally, we summarize the integer linear program to be solved as:
\begin{eqnarray}
\phantom{xxxx} & \min\limits_{\mf{y}}\ & \mf{c}_R^T\, \mf{y}_R\ +\ \mf{c}_B^T\, \mf{y}_B
\label{eq:len_ilp}\\
& \mbox{s.t.} & \sum\limits_{f\in \mathcal{F}} m_e^f\, y_R^f\ =\ 
\sum\limits_{l\in \mathcal{E}^O} m_e^l\, y_B^l\ \ \ \forall e \in \mathcal{E}\nonumber\\
&& y_R^f \in \{0,1\}\ \  \forall f \in \mathcal{F}\nonumber\\[1mm]
&& y_B^l \in \{0,1\}\ \  \forall l \in \mathcal{E}^O\ \ \ . \nonumber
\end{eqnarray}
As we will show now this problem can be efficiently solved by
computing a graph min-cut.

\subsection{Relation to Graph Cuts}

Discrete approaches to image segmentation are very well studied in
computer vision and the vast majority uses pixels as their basic
regions. However, they do usually not express (length) regularity in
terms of the boundary segments of these regions - this would imply a
four-connectivity.

Instead, these approaches are based on \emph{graphs} where the basic
regions correspond to nodes and the length term is represented in
terms of \emph{edges} that connect pairs of nodes
\cite{Boykov-Kolmogorov-03}. We denote the set of nodes $\mathcal{P}$,
where each $p \in \mathcal{P}$ represents the center point of a basic
region $f\in \mathcal{F}$. Analogous to the above integer program,
each center $p$ of a basic region is associated a binary variable $y_p
\in \{0,1\}$ indicating foreground and background.  The smoothness
term is modeled by a set of edges $\mathcal{N}$ (also called
neighborhood) and for length regularity can be expressed as:
\begin{eqnarray*}
\phantom{xxxx} |C| & \approx & \frac{1}{k(\mathcal{N})}
\sum\limits_{(p,q)\in \mathcal{N}} \frac{1}{\|p-q\|} (1-\delta(y_p,y_q))\ \ \ ,
\end{eqnarray*}
where $k(\mathcal{N})$ is a normalization constant that ensures that
the weights remain comparable when the size of the neighborhood is enlarged.

Minimizing this energy can be written as a \emph{minimum cut} problem,
which again can be written as 
$$
\phantom{xxxx} \min\limits_{\mf{y}_P\in \{0,1\}^{|\mathcal{P}|}}\, \mf{c}_R^T\, \mf{y}_P \ +\ \sum\limits_{(p,q)\in 
\mathcal{N}} w_{p,q}\, |y_p - y_q| \ \ .
$$
It is well-known, e.g.\ \cite{Dantzig-Thapa-97}, that such absolutes can
be rewritten as linear programs, i.e.\ this problem can be equivalently written
as
\begin{eqnarray*}
\phantom{xxx} &\min\limits_{\mf{y}_P,\mf{a}_\pm}\ & \mf{c}_R^T\, \mf{y}_P \ +\ \sum\limits_{(p,q)\in 
\mathcal{N}} w_{p,q}\, (a_{p,q}^+ + a_{p,q}^-)\\
& \mbox{s.t.} & x_p - x_q\, =\, a_{p,q}^+ - a_{p,q}^{-}\ \ \forall (p,q)\in \mathcal{N}\\[1mm]
& & \mf{y}_P\in \{0,1\}^{|\mathcal{P}|}\, ,\ \   a_{p,q}^+, a_{p,q}^- \ge 0 \ \ \forall (p,q)\in \mathcal{N}\ .
\end{eqnarray*}
If we assume that the neighborhood links exactly all pairs of
neighboring basic regions (i.e.\ those that share a boundary segment),
these are exactly our surface continuation constraints
\eqref{eq:len_surf_cont}.  Since graph cuts can be optimized globally
efficiently it follows that \eqref{eq:len_ilp} is polynomial-time
solvable.

In summary, what we have proposed so far is a restriction of graph
cuts to graphs that are planar when source and sink are removed. This
restriction is crucial for (our solution of) the problem we really
want to solve: to integrate curvature regularity into the framework.

\begin{figure}
\setlength{\tabcolsep}{1.5mm}
\begin{tabular}{cc}
\includegraphics[width=0.475\linewidth]{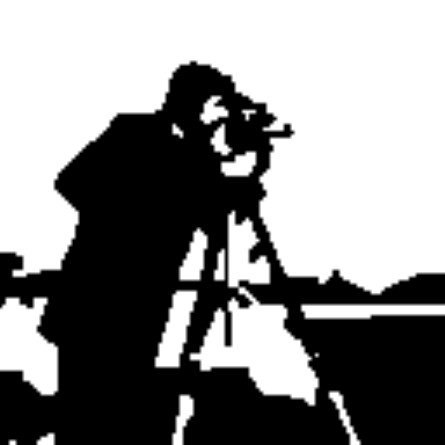} &
\includegraphics[width=0.475\linewidth]{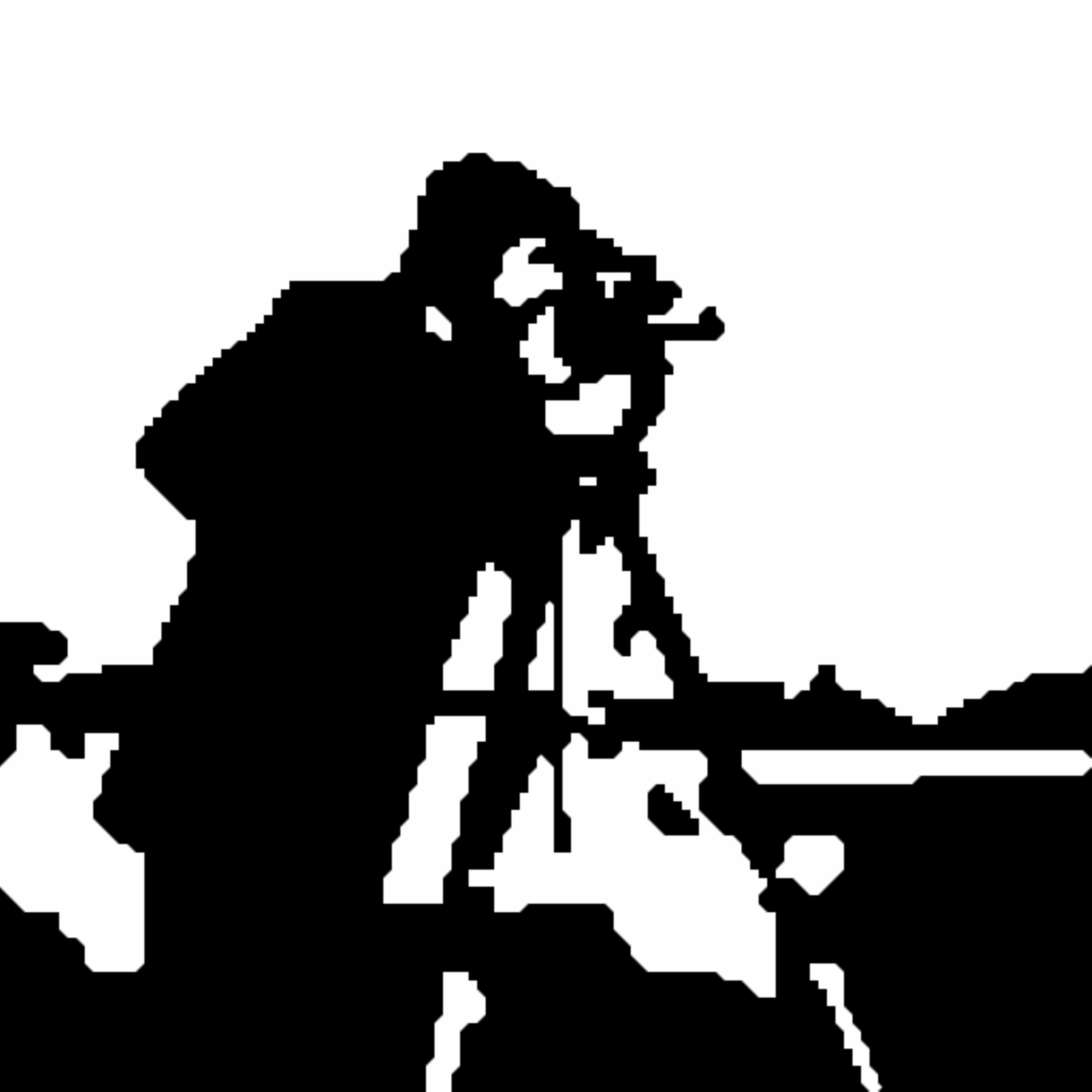}\\[-1mm]
{\small Standard Graph Cut} & {\small Subdivision}
\end{tabular}
\caption{Comparison of standard graph cuts and the novel subdivision scheme, both run
with the same length weight and an 8-connectivity. (Input image shown
in Figure \ref{fig:curv_seg_exp}.)}
\label{fig:gc_comp}
\end{figure}

In contrast to standard applications of graph cuts our method relies
on subdiving pixels. In the case of a very strong data term a boundary
pixel will probably not be split into a foreground and a background
part. Standard graph cuts might then be the better way to reflect the
length regularity term. However, in practice boundary pixels usually
have weak data terms due to partial overlap. Figure \ref{fig:gc_comp}
shows the resulting segmentations of both schemes, where the input
image can be found in Figure \ref{fig:curv_seg_exp}. Here it can be
seen that the novel scheme gives access to a higher resolution and
produces slightly different segmentations. These are actually of lower
energy than for the standard scheme. Hence, if anything we have
\emph{gained} something with the novel formulation.

\subsection{Relation to Sullivan's Method}

Sullivan \cite{Sullivan-PhD-92,Sullivan-94} gave a method for finding
a surface of minimal (weighted) area in 3D-space\footnote{In fact, he
considered general $N$-D spaces, but this is not important here.} 
subject to the constraint that the surface span a given (set of)
boundary lines. His method also relies on a cell complex and can be
written as a polynomial-time solvable minimum cost flow
problem. Further, the method allows integrating volume terms.

In essence, what we have done so far is a restriction of this method
to 2D and where the set of prescribed boundary elements is empty
(these objects would be one-dimensional). However, there is one major
difference: in Sullivan's method all variables are unrestricted,
whereas we have the constraints $y_R^f \in \{0,1\}$ and $y_B^l \in
\{0,1\}$. And of course our main goal is to integrate curvature
regularity.

\section{Handling Curvature Regularity}

We now show how the described integer linear program can be
generalized to curvature regularity, i.e.\ to the model
\begin{eqnarray}
\phantom{xxxx} & \min\limits_{u: \Omega \rightarrow \{0,1\}}&\
\int\limits_\Omega\! g(\mf{x})\, u(\mf{x})\, d\mf{x} + \nu\, |C|\nonumber \\ 
&& + \lambda\! \int\limits_C  |\kappa_C(\mf{x})|^p d\mathcal{H}^1(\mf{x})\ \ \ .
\end{eqnarray}
Here $\lambda > 0$ is a curvature weight, $\kappa_C(\mf{x})$ stands
for the curvature of the line $C$ at a given point $\mf{x}$ of the
line and $p > 0$ is an arbitrary exponent (usually $p=2$ is a good
choice). The notation $d\mathcal{H}^1(\mf{x})$ signifies that the
integral is over a set of lines and that it is independent of the
parameterization of these lines.

We emphasize that our method allows more general regularity terms,
namely arbitrary positive functions depending on position, direction
and absolute curvature. In particular this allows spatially weighted
regularity terms.

\subsection{Discretizing the Problem}

Again, our solution is based on a cell-complex and the data term is
handled in the exact same way as above. That is, we again have region
indicator variables $x_f$ for all $f\in \mathcal{F}$. We were able to
express the length regularity in terms of (single) boundary
segments. For curvature, this is not possible: all boundary segments
are straight lines, hence have curvature $0$ everywhere. The only
points where non-zero curvature can occur are the meeting points of
two boundary segments. Hence it is common to consider pairs of line
segments \cite{Parent-Zucker-89,Amini-et-al-90} to express curvature
regularity. So far, however, this was not compatible with region
terms.

For every pair $l_1,l_2$ of adjacent line segments with compatible
orientations we now have an indicator variable $\mf{y}_B^{l_1,l_2} \in
\{0,1\}$. Here we follow the convention that the line segment that is 
traversed earlier is also listed first in the pair. We get a cost
function of the form
\begin{equation}
\phantom{xxxx} \mf{c}_R^T\, \mf{y}_R\ +\ \mf{c}_B^T\, \mf{y}_B \ \ ,
\label{eq:curv_ilp_cost}
\end{equation}
where $\mf{y}_B$ now contains all the pairwise variables. We proceed
to describe the entries of the corresponding cost vector $\mf{c}_B^T$.

\subsection{Computing the Weights}

\label{sec:weights}

\begin{figure}
\begin{center}
\includegraphics[width=0.25\textwidth]{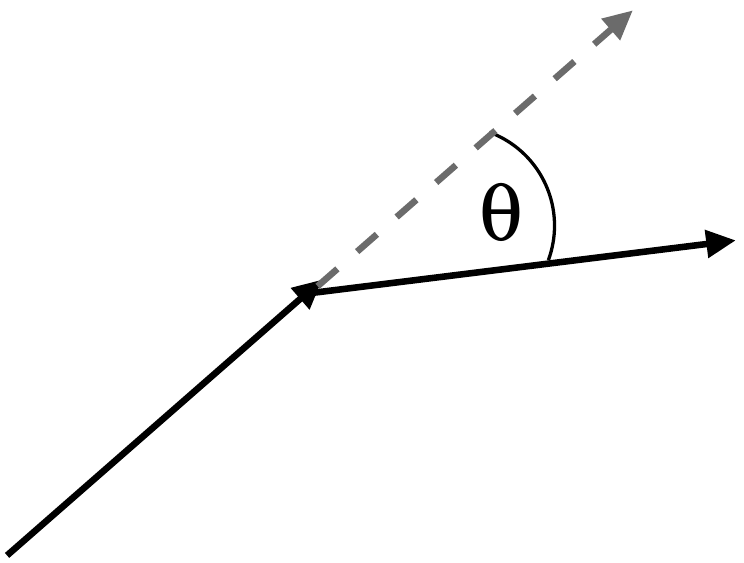}
\end{center}
\caption{The angle $\theta$ is the basis for computing the curvature of the pair of black lines.}
\label{fig:theta}
\end{figure}

Computing curvature from two adjacent line segments is based on
considering the direction change, measured by the angle $\theta$ in
Figure \ref{fig:theta}.

There are basically two ways to compute the term $|\kappa|^p$ from
this angle: firstly, we can just take the power $p$ of the angle
(which should be measured in arc length). The second method is based
on the work of Bruckstein et al.\ \cite{Bruckstein-et-al-01}, and
makes also use of the lengths $\ell(l_1)$ and $\ell(l_2)$ of the two
lines: $$
\phantom{xxxx} \min\{\ell(l_1),\ell(l_2)\} \left( \frac{\theta}{\min\{\ell(l_1),\ell(l_2)\}  } \right)^p\ \ .
$$ The original idea of Bruckstein et al.\ was to take the longest
straight lines pre- and succeeding a direction change. In our context
we can only consider elementary boundary segments, so we do not have
the same convergence properties. In practice we found that both
weights work fine and Bruckstein et al's weights significantly reduce
the running time of the employed linear programming solver.

Denoting either of the two arising weights as $w_{l_1,l_2}$, we get
one of two components for the cost entry $c_{l_1,l_2}$. Together with
length regularity this entry is $$
\phantom{xxxx} c_{l_1,l_2} = w_{l_1,l_2}\, +\, \tfrac{1}{2} \big(\ell(l_1) + \ell(l_2)\big)\ \ .
$$ There are however some special cases involving the image border, a
point that we neglected in \cite{Schoenemann-Kahl-Cremers-09}:
firstly, at the four corners of a rectangular domain $\Omega$ we will
want to set the curvature weight to $0$. Note that we still penalize
the angle in which a region boundary meets the domain boundary
$\partial \Omega$.

A second point relates to the length weight: whenever one (or two) of
the line segments in a pair is part of the domain border, its length
should be set to $0$. Otherwise there would be a bias towards
associating the regions at the image border to the background.

\subsection{An Adequate Constraint System}

\label{sec:curv_constraints}

As before for length regularity, the linear cost function
\eqref{eq:curv_ilp_cost} needs to be minimized subject to suitable
constraints to closely reflect the discretized model function.  In
\cite{Schoenemann-Kahl-Cremers-09} we presented two sets of
constraints, with the aim to ensure that the boundary variables indeed
describe a boundary of the region variables. In this work we show that
two more sets of constraints are needed, where the second one becomes
necessary only when several regions meet in a single point. In
particular, we will show that in this case there are several valid
boundaries and that our method searches for the one with least cost.

\begin{figure}
\begin{center}
\begin{tabular}{cc}
\includegraphics[width = 0.14\textwidth]{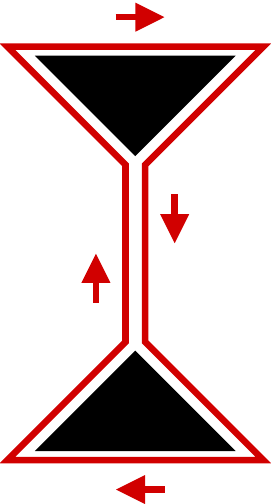} &
\includegraphics[width = 0.14\textwidth]{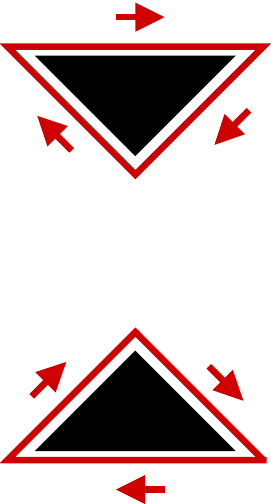}\\
{\small (a)} & {\small (b)}
\end{tabular}
\end{center}
\caption{Without the boundary consistency constraints the configuration 
in (a) is valid and -- for squared curvature and without length
penalty-- cheaper than the desired one in (b). With boundary
consistency (b) remains feasible, but (a) is excluded, as desired.  }
\label{fig:touching_lines}
\end{figure}

Firstly, we adapt the surface continuation constraints to the new kind
of boundary variables. To this end, we define the incidence
$m_e^{l_1,l_2}$ of a line segment pair and a boundary segment
$e\in\mathcal{E}$. This value is defined as $0$ unless $l_1$ is an
orientation of $e$ (note that $l_2$ is irrelevant for this value).
Otherwise the value is the same as the previously defined incidence of
$l_1$ and $e$. Hence, the surface continuation constraints read
\begin{equation}
\phantom{xxxx} \sum\limits_{f\in \mathcal{F}} m_e^f\, y_R^f\ =\ 
\sum\limits_{l_1, l_2 \in \mathcal{E}^O} m_e^{l_1,l_2}\, y_B^{l_1,l_2}\ \ \ \forall e \in \mathcal{E} \ \ .
\end{equation}
These constraints alone leave a lot of freedom. In particular, we
could choose line pairs without direction changes everywhere. What is
really wanted, though, is that for every active $y_b^{l_1,l_2}$ both
$l_1$ and $l_2$ belong to the region boundary induced by the region
variables. This is ensured by two sets of constraints, where the first
is called \emph{boundary continuation}. In words it can be stated as
\begin{quote}
\textbf{Boundary Continuation Constraint:}
\textit{
If a pair of line segments $l_1,l_2$ is active, there must be a
succeeding pair $l_2,l_3$ that is also active. Likewise there must be
a preceeding active pair $l_0,l_1$.}
\end{quote}
These constraints ensure that the active line pairs actually define
closed paths. They are identical to the constraints arising for the
computation of shortest paths in a graph such as \cite{Amini-et-al-90}
and are stated as $$
\sum\limits_{l_0} y_B^{l_0,l_1} = \sum\limits_{l_2} y_B^{l_1,l_2}
\ \ \ \forall\, l_1 \in \mathcal{E}^O\ \ .  $$ 
Now we have paths, but we cannot guarantee that all parts of these
paths are actually region boundaries. Indeed, we invite the reader to
check that the configuration in Figure \ref{fig:touching_lines} (a)
satisfies all constraints introduced so far. Moreover, for small
length weights and squared curvature the cost of this configuration
will be lower than those of the desired configuration shown in part
(b).
To exclude cases such as (a) from the optimization, we add a new set of constraints:
\begin{quote}
\textbf{Boundary Consistency Constraint:}
\textit{
For every boundary segment, only one of the two possible orientations
can be active.  }
\end{quote}
To formalize this, we denote by $e^\rightarrow$ and $e^\leftarrow$
the positive and negative orientation of a boundary segment $e$. Then
the constraint can be written as
$$
\sum\limits_{l_1} y_B^{l_1,e^\leftarrow} +\, \sum\limits_{l_2} y_B^{e^\rightarrow,l_2} \le 1 \ \ \forall e \in \mathcal{E}\ \ .
$$ Similar sets of constraints can be derived, e.g.\ $\sum_{l_1}
y_B^{e^\leftarrow,l_1} +\, \sum_{l_2} y_B^{l_2,e^\rightarrow}\le 1$,
but experimentally we found them to be redundant. Moreover, if $e$ is
part of the domain border $\partial \Omega$ then one of the
orientations will never occur in a desired configuration. We simply
disregard the corresponding variables.

\begin{figure}
\begin{center}
\begin{tabular}{ccc}
\includegraphics[height=3.5cm]{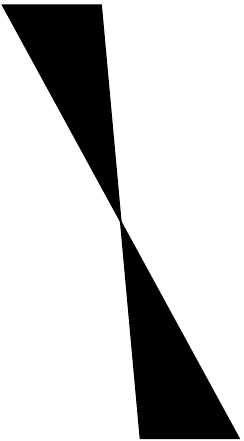} &
\includegraphics[height=3.5cm]{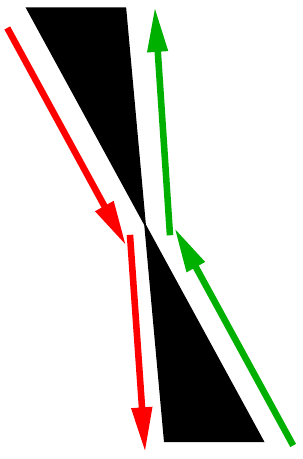} &
\includegraphics[height=3.5cm]{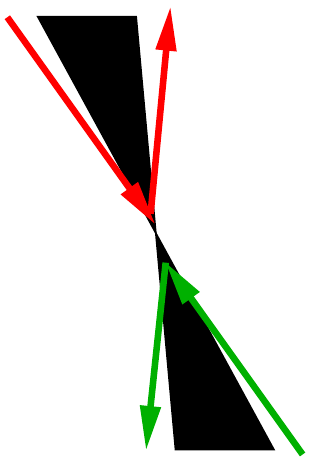} \\[-1mm]
{\small (a)} & {\small (b)} & {\small (c)}
\end{tabular}
\end{center}
\caption{Determining boundaries is ambiguous: (a) A segmentation where
two regions meet in a point. (b) A low-cost boundary configuration,
where different pairs are indicated by different colors. (c) An
equally valid, but more expensive boundary.}
\label{fig:ambig}
\end{figure}

\begin{figure}
\begin{center}

\includegraphics[height=4.5cm]{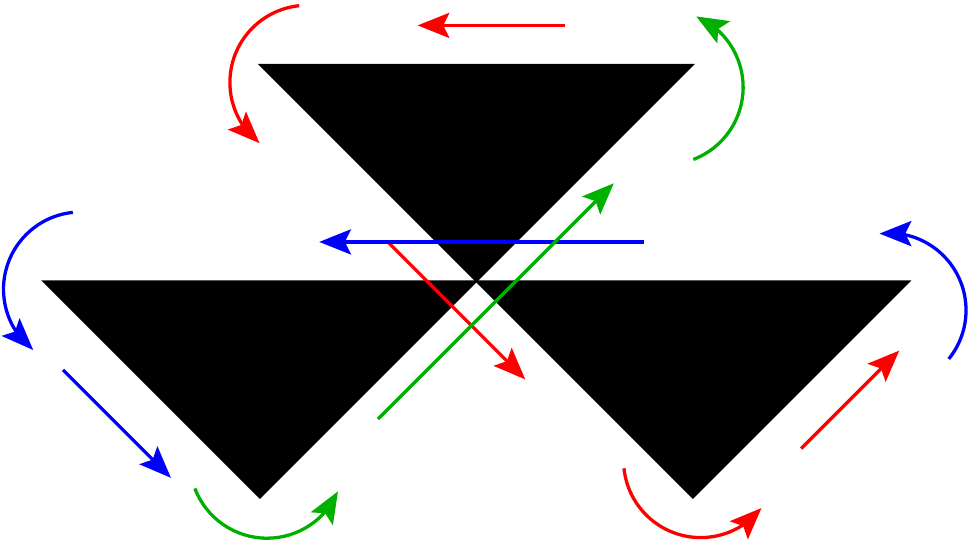}

\caption{If three or more regions meet in a point, self-intersecting
boundaries can define valid segmentations. If these are undesired, a
fourth constraint set is needed.  The different colors denote
different phases when traversing the line. There is only one line.}
\label{fig:three_regions}
\end{center}
\end{figure}

With these constraints, there is still one issue left to take care of,
and it affects cases where several regions meet in a point. In this
case, as shown in Figure \ref{fig:ambig}, there are several valid
boundary configurations.  As long as only two regions meet, the given
constraint system ensures that indeed the configuration with lower
cost is selected. However, as soon as there are three or more regions
meeting in a point, this constraint system will allow configurations
\emph{with crossings}, as exemplified in
Figure~\ref{fig:three_regions}. Should we really avoid such
configurations? There is actually no obvious answer if one has in mind
that, in the theory of continuous plane curves, the set enclosed by a
positively oriented closed curve can be defined as the points of
positive {\it index} (also called {\it winding
number})~\cite{Rudin:1987}. In the example of
Figure~\ref{fig:three_regions}, the index of each point in a black
triangle with respect to the outer curve defined by the arrows is one,
so it makes sense to consider the curve as a region
boundary~\footnote{the index of a point on a discrete grid with
respect to a discrete curve can also be computed, see for instance the
winding number algorithm at
www.softsurfer.com/Archive/algorithm\_0103/algorithm\_0103.htm}. For
the sake of comparison and completeness, we provide in
section~\ref{sec:experiments}, Figure~\ref{fig:cross_prev} a couple of
experiments done either with or without crossing prevention. In the
former case, we add a new set of constraints:
\begin{quote}
\textbf{Crossing Prevention Constraint:}
\textit{
If two pairs of line segments cross, only one of them may be active.
}
\end{quote}

Denoting $\mathcal{C}$ the set of crossing line pairs, this constraint
is easily formalized as
$$
\phantom{xxxx} y_B^{l_1,l_2}\, +\, y_B^{l_3,l_4} \le 1\ \ \ \forall (l_1,l_2,l_3,l_4) \in \mathcal{C}\ \ .
$$ In practice these constraints are ignored in a first
phase. Afterwards, the (usually very few) violated constraints are
added in passes and the system is re-solved, until there are no more
violated constraints. In our experiments we never needed more than $9$
passes. However, solving the arising programs can be quite time
consuming, even when starting from the previous configuration.

In summary, region-based segmentation with curvature regularity is now
expressed as the integer linear program
\begin{eqnarray*}
&\min\limits_{\mf{y}_R,\mf{y}_B}& 
 \mf{c}_R^T\, \mf{y}_R\ +\ \mf{c}_B^T\, \mf{y}_B\\
&\mbox{s.t.} & \sum\limits_{f\in \mathcal{F}} m_e^f\, y_R^f\ =\ 
\sum\limits_{l_1, l_2 \in \mathcal{E}^O} m_e^{l_1,l_2}\, y_B^{l_1,l_2}\ \ \ \forall e \in \mathcal{E}\\
&& \sum\limits_{l_0} y_B^{l_0,l_1} = \sum\limits_{l_2} y_B^{l_1,l_2}
  \ \ \ \forall l_1 \in \mathcal{E}^O\\
&& \sum\limits_{l_1} y_B^{l_1,e^\leftarrow} +\, \sum\limits_{l_2} y_B^{e^\rightarrow,l_2} \le 1
\ \ \ \forall e \in \mathcal{E}\\
&&y_R^f \in \{0,1\} \ \forall f\in\mathcal{F},\ \, y_B^{l_1,l_2} \in \{0,1\}\ \forall\, l_1,l_2 \in \mathcal{E}^O\, ,
\end{eqnarray*}
and the optional constraints
$$
y_B^{l_1,l_2}\, +\, y_B^{l_3,l_4} \le 1\ \ \ \forall (l_1,l_2,l_3,l_4) \in \mathcal{C}\ .
$$

A first evaluation of this new scheme was given in Figure
\ref{fig:intro} on page \pageref{fig:intro} which shows that curvature
regularity is better suited to ensure connected regions in the
presence of long and thin objects. Note that the found solutions for
curvature are generally not globally optimal. Details on the
optimization scheme are given in Section~\ref{sec:optimization} and
more experiments will be given in Section~\ref{sec:experiments}.
First, however, we discuss how to handle the problem of image
inpainting.

\section{Inpainting}

In image inpainting we are given an image $I : \Omega \rightarrow \R$
together with a damaged region $\Omega_d \subset \Omega$. This damaged
region can have arbitrarily many connected components and each of
these can enclose holes. The task is to fill the damaged region with
values that fit nicely with the values of $I$ outside the damaged
region.

To this end, the above integer linear program is generalized to 
give a \emph{structured} inpainting approach, where we consider
the continuous model \cite{Masnou-Morel-98,Masnou-02,Chan-Kang-Shen-02}
\begin{eqnarray}
&\min\limits_{u: \Omega \rightarrow \{0,1\}}& \int\limits_{I_l}^{I_u}  
\int\limits_{C_{u,t}} |\kappa_{C_{u,t}}(\mf{x})|^p\,  
d\mathcal{H}^1(\mf{x}) \, dt\label{eq:inp_model}\\
&& \mbox{s.t. } u(\mf{x}) = I(\mf{x})\ \ \forall \mf{x} \in \Omega \setminus \Omega_d\ \ ,
\nonumber
\end{eqnarray}
where $I_l$ and $I_u$ are the minimal and maximal intensities of $I$
along the border of the damaged region, $C_{u,t} = \{ \mf{x}\, |\,
u(\mf{x}) = t \}$ is the set of level lines for level $t$ of
$u(\cdot)$ and again $p > 0$ is an exponent for curvature.

For the case of absolute curvature ($p=1$) and that $\Omega_d$
consists of \emph{singly-}connected components (i.e.\ the components
do not enclose holes), an efficient global optimization scheme was
given in \cite{Masnou-Morel-98,Masnou-02}. We are interested in the
more general problem of arbitrary domains and exponents $p > 0$. In
particular, $p=2$ is usually a better model.

\subsection{Discretization}

As before for image segmentation, our strategy is to discretize the
model \eqref{eq:inp_model} by introducing basic regions and pairs of
boundary segments. However, the regions are no longer limited to two
labels: the intensity $u_f$ of region $f \in \mathcal{F}$ can be
anywhere between $I_l$ and $I_u$. We follow the strategy in
\cite{Masnou-Morel-98,Masnou-02} and consider only integral values inside
this range. The result is a fully discrete labeling problem.

Naturally, one also needs to change the right-hand-side values of the
inequality constraints. Moreover, for inpainting we always include the
crossing prevention constraints since by definition level lines cannot
cross.

To make sure that the boundary variables truly reflect level lines, it
would be advisable to associate each basic region multiple variables,
reflecting level sets. Then, there would be a \emph{binary} variable
$y_f^k$ for every integral value $I_l \le k \le I_u$ where a value of
$1$ reflects that the intensity $u_f$ of the basic region is at least
equal to $k$ (i.e.\ $u_f \ge k$). For $k' > k$ this would naturally
entail the constraints $y_f^{k'} \ge y_f^k$. Moreover, there would be
binary variables $y_{l_1,l_2}^k$ that would be forced to be consistent
with the level variables exactly as for binary segmentation.

This strategy is however not practicable for the domain sizes we want
to address: the problem is way too large scale. Instead, we settle for
one integral variable $y_R^f\in \{I_l,\ldots,I_u\}$ for every region
$f\in\mathcal{F}$, directly reflecting the intensity $u_f$ of the
face. In addition, there are boundary variables reflecting the
intensity differences between neighboring regions. Inside the damaged
domain they can be restricted to values $y_B^{l_1,l_2}\in [ 0, I_u -
I_l ]$. For the fixed part, we only consider basic regions that border
on the damaged region. At their other borders the respective boundary
variables can take values in $\{0,\ldots,I_u\}$. In practice it is
advisable to first subtract the constant $I_l$ from the entire image
(note that each connected component of $\Omega_d$ can be processed
independently).

\begin{figure}
\begin{center}
\begin{tabular}{cc}
\includegraphics[width=0.4\linewidth]{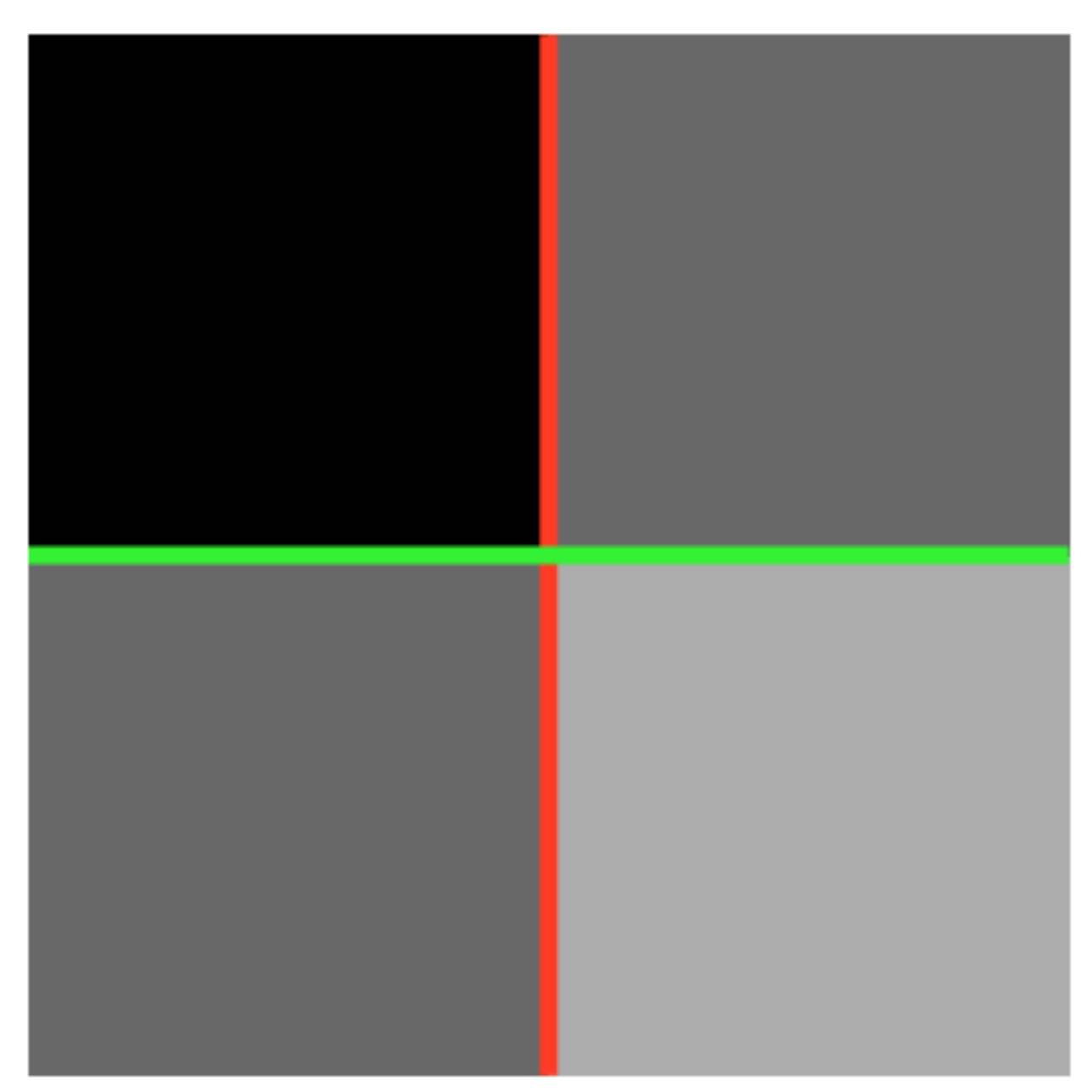} &
\includegraphics[width=0.4\linewidth]{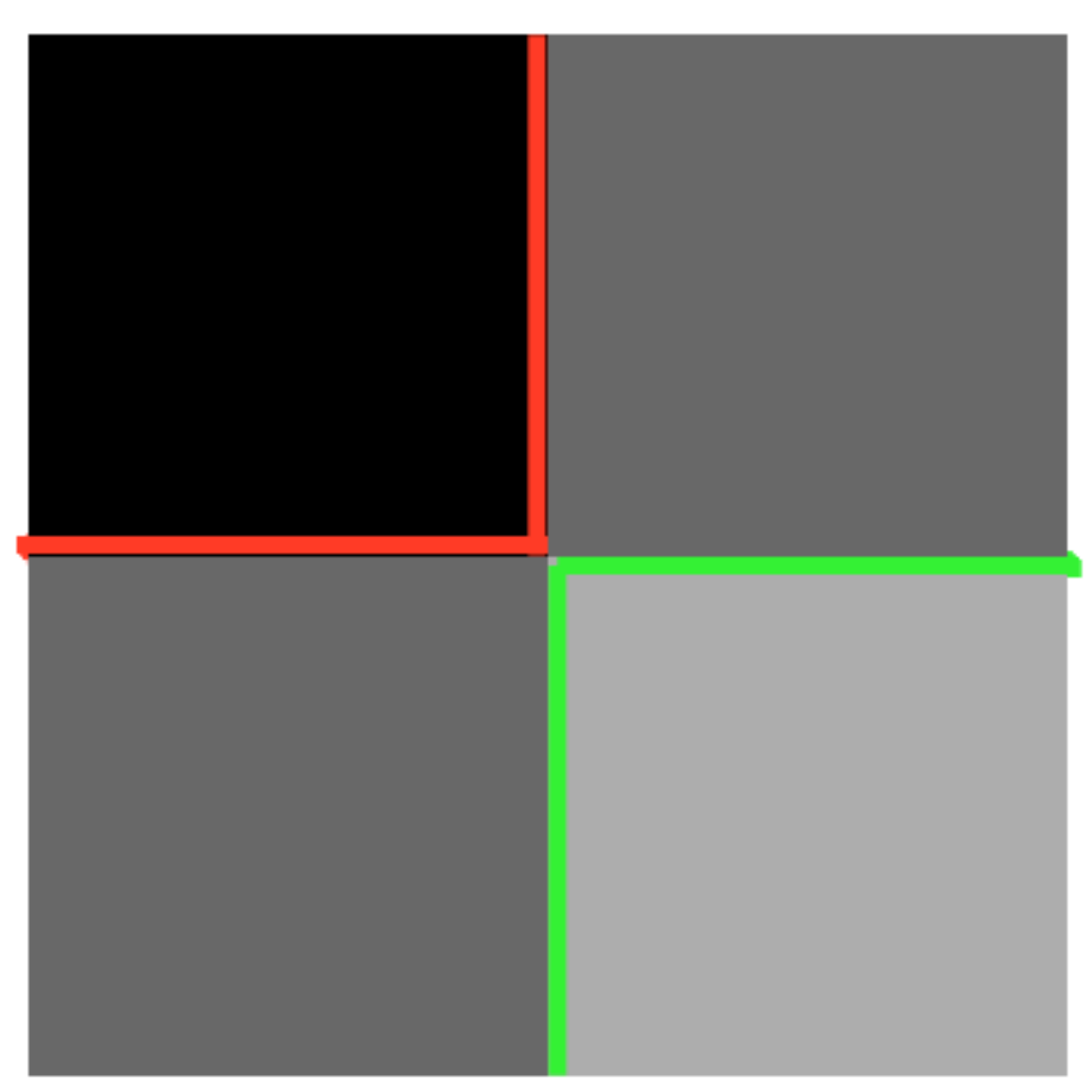}\\
{\small (a)} & {\small (b)}
\end{tabular}
\end{center}
\caption{Best viewed in color. Estimated boundary variables for 
the given intensity profiles. Colors denote different pairs (not all
pairs are shown) (a) With the proposed approximation we do not always
get level lines. Here the boundary variables may jump between
different intensity levels. (b) With level variables one would get the
correct solution, but the computational demands are too high in
practice. }
\label{fig:level_lines}
\end{figure}

As shown\footnote{Many thanks to Yubin Kuang for providing these
images.} in Figure \ref{fig:level_lines} this strategy does generally
not give level lines, but it is a reasonable approximation. The
arising integer linear program is stated as
\begin{eqnarray}
\phantom{xx} & \min\limits_{\mf{y}_R,\mf{y}_B} &\ \mf{c}_B^T \mf{y}_B \\
& \mbox{s.t.} & \sum\limits_{f\in \mathcal{F}} m_e^f\, y_R^f\ =
\sum\limits_{l_1, l_2 \in \mathcal{E}^O} m_e^{l_1,l_2}\, y_B^{l_1,l_2}\ 
\ \ \forall e \in \mathcal{E}\nonumber \\
&& \sum\limits_{l_0} y_B^{l_0,l_1} = \sum\limits_{l_2} y_B^{l_1,l_2}
  \ \ \ \forall l_1 \in \mathcal{E}^O\nonumber \\
&& \sum\limits_{l_1} y_B^{l_1,e^\leftarrow} +\, \sum\limits_{l_2} y_B^{e^\rightarrow,l_2} \le I_u
\ \ \ \forall e \in \mathcal{E}\  \nonumber\\
&& y_B^{l_1,l_2}\, +\, y_B^{l_3,l_4} \le I_u\ \ \ \forall (l_1,l_2,l_3,l_4) \in \mathcal{C}\nonumber \\
&& y_R^f = I_f - I_l\ \ \ \forall\, f \subset \Omega \setminus \Omega_d\nonumber\\[1mm]
&& y_R^f \in \{0,\ldots,I_u-I_l\}\ \ \ \ \forall\, f \subset \Omega_d\nonumber\\[1mm]
&& y_B^{l_1,l_2} \in \{0,\ldots,I_u-I_l\}\ \ \ \ \forall\, l_1,l_2 \in \mathcal{E}^O\ \ .
\nonumber
\end{eqnarray}
Note that we have one such program for every connected component of
$\Omega_d$.

\subsection{Estimating Incoming Level Lines}

A correct estimation of the direction of the level lines touching from
outside the inpainting domain $\Omega$ is important, since the method
aims at prolonging them with no additional curvature, if possible. We
follow the simple and efficient method proposed by Bornemann and
M{\"a}rz in~\cite{Bornemann-Maerz-07} after the work of
Weickert~\cite{Weickert-Book-98,Weickert-03} on the robust
determination of coherence directions in image: at a point
$x\in\Omega\setminus
\Omega_d$, the coherence direction is the normalized eigenvector
associated to the minimal eigenvalue of the structure
tensor~\cite{Bornemann-Maerz-07}:
\begin{equation}
\phantom{xxxx} J(x)=\frac{\Bigl(K_\rho\ast\Bigl(\one{\Omega\setminus \Omega_d}\nabla I_{\sigma}\otimes 
\nabla I_{\sigma}\Bigr)\Bigr)(x)}{\Bigl(K_\rho\ast\one{\Omega\setminus \Omega_d}\Bigr)(x)}
\end{equation}
where $\one{\Omega\setminus \Omega_d}$ is the characteristic function
of $\Omega\setminus \Omega_d$, $I_{\sigma}$ is defined as
\begin{equation}
\phantom{xxxx}I_{\sigma}=\frac{K_\sigma\ast\Bigl(\one{\Omega\setminus \Omega_d}I\Bigr)}
{K_\sigma\ast\one{\Omega\setminus \Omega_d}},
\end{equation}
and $K_\rho$, $K_\sigma$ are Gaussian smoothing kernels with standard
deviations $\rho$ and $\sigma$. Experimentally, setting $\sigma=1.5$,
$\rho=4$ yields a reliable estimation of the incoming level lines
directions along $\partial\Omega_d$.

\section{Optimization Strategies}

\label{sec:optimization}

In general, solving integer linear programs is an NP-hard problem
\cite{Schrijver-Book-86}.  In some cases, where there are linear
\emph{in}equality constraints $\mf{A}\mf{x} \le \mf{b}$ and the matrix
$\mf{A}$ is \emph{totally unimodular} one can find the global optimum
by solving the linear programming relaxation \cite{Schrijver-Book-86},
i.e.\ the problem one obtains when dropping all integrality
constraints on the variables. For instance, a constraint $y_i \in
\{0,1\}$ will be relaxed to $y_i \in [0,1]$. The arising problem can
be solved in (weakly) polynomial time using interior point methods
\cite{Ye-Book-97}.

Several of the discussed systems are in fact polynomial-time solvable,
in particular the length-based problems and the boundary continuation
constraints by themselves. The integer linear program for curvature
regularity is however not in this class. Still, experimentally we
found that solving the linear programming relaxation often gives
nearly integral solutions. Moreover, the relaxation value provides a
(usually quite tight) lower bound on the original problem. We proceed
to discuss this strategy in detail.

\subsection{Solving the Linear Programming Relaxation}

There are two popular ways of solving general linear programs.
Firstly, there is the dual simplex method \cite{Dantzig-Thapa-97},
based on refactorizing the constraint system. It is usually the most
memory saving method and in practice superior to the primal simplex
method. For some variants of the simplex method exponential worst-case
run-times have been proved. On practical problems the method often
works very well, and there are competitive and freely available
implementations, e.g.\ the solver
Clp\footnote{\url{http://www.coin-or.org/projects/Clp.xml}}. We found
it quite useful for an 8-connectivity, but for a 16-connectivity --
and very low length weights -- we got acceptable running times only
for some of the images we tried. In other cases the solver was
terminated after several days without having solved the problem. In
the end we chose the length weights high enough to get acceptable
running times. The results we get indicate that these settings
actually provide a very good model.

On the other hand, there are interior point methods which usually
perform Newton-iterations on a primal-dual formulation of the problem.
This entails frequent matrix inversions, solved via the sparse
Cholesky decomposition. We are not aware of a freely available solver
that performs well on large scale problems such as ours. Hence, we
tested several commercial packages and found Gurobi and FICO Xpress to
be well-suited solvers for our problem. Generally the running times of
commercial interior point solvers are quite predictable. The downside
is the memory consumption: we found that a little more than twice the
memory consumption of a simplex solver is needed for our problem,
where we tested with an 8- and a 16-connectivity.

For segmentation, the combination of licensing issues and the high
memory demands made us use the simplex method. For inpainting, where
the memory demands are lower, the interior point methods proved
superior.


\subsection{Obtaining and Evaluating Integral Solutions}

\label{sec:integral_solutions}

Solving the linear programming relaxation provides a fractional
solution as well as a lower bound on the original integral problem.
Since the fractional solutions are often close to integral, we derive
an integral solution by simply thresholding the region variables. This
already defines a segmentation, but we would also like to know its
energy, so we have to infer the associated boundary variables.

We already discussed the complications in case two or more regions
meet in a point (see Section~\ref{sec:curv_constraints}). Our method
will then select the cheapest allowed boundary configuration according
to the selected constraint set. In contrast, the recent method of
El-Zehiry and Grady \cite{ElZehiry-Grady-10} will take the sum of all
configurations here.

Due to these difficulties we so far did not implement a specialized
routine to compute the energy of a segmentation, although this should
be possible. Instead, we simply re-run the linear programming solver,
this time with all region variables fixed according to the
segmentation. Note that this strategy was also pursued in
\cite{Schoenemann-Kahl-Cremers-09}, so the gaps reported there are w.r.t.
the integer program, not the model itself.

In case crossing prevention was selected we again add violated
constraints in passes. In addition, we fix all ``impossible'' boundary
variables to $0$, where impossible refers to pairs of line segments
where along one of the line segments the segmentation stays constant.
In the vast majority of cases this produced an integral solution and
hence the optimal boundary configuration. In a few cases we got up to
$12$ fractional variables. We presently assume that the computed cost
is close enough to the actual cost.

\section{Experiments}

\label{sec:experiments}

In this section we evaluate the proposed scheme for both image
segmentation and inpainting, where in all cases we consider the
intensity range $[0,255]$. For image segmentation we also evaluate how
close to the global optimum we got. The experiments were run on a 3.0
GHz Core2 Duo machine equipped with $8$ GB of memory. For segmentation
we used the dual simplex method in the solver Clp, for inpainting we
used the interior point method of Gurobi.

\subsection{Image Segmentation}

For binary image segmentation, we show experiments for a totally
unsupervised problem and an interactive one where seed nodes are
given.

\paragraph{Unsupervised Image Segmentation}

For unsupervised image segmentation we use data functions $g_0, g_1$
as in the piecewise constant functional of Mumford and Shah
\cite{Mumford-Shah-89}, i.e. where the intensity of an image point is compared
to the mean value of a region. This results in the model:
\begin{eqnarray}
&& \int\limits_\Omega \!\!
\Big( I(\mf{x}) - \mu_0 \Big)^{\!2} [1 - u(\mf{x})]\, d\mf{x}\, + 
\int\limits_\Omega\!\! \Big( I(\mf{x}) - \mu_1 \Big)^{\!2} u(\mf{x})\, d\mf{x}
\nonumber\\
&& \hspace{1cm} +\, \nu\, |C| + \lambda\! \int\limits_C  |\kappa_C(\mf{x})|^2\, d\mathcal{H}^1(\mf{x}) \ \ \ . \label{eq:unsupervised}
\end{eqnarray}
We set the mean values $\mu_0,\mu_1$ to the minimal and maximal
intensity in the given image, respectively.

\begin{figure*}
\begin{tabular}{ccccc}
\includegraphics[width=4cm]{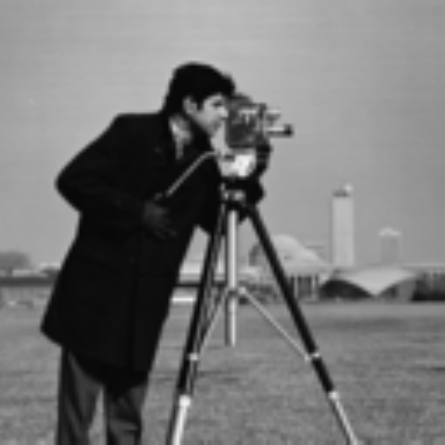} &
\includegraphics[width=4cm]{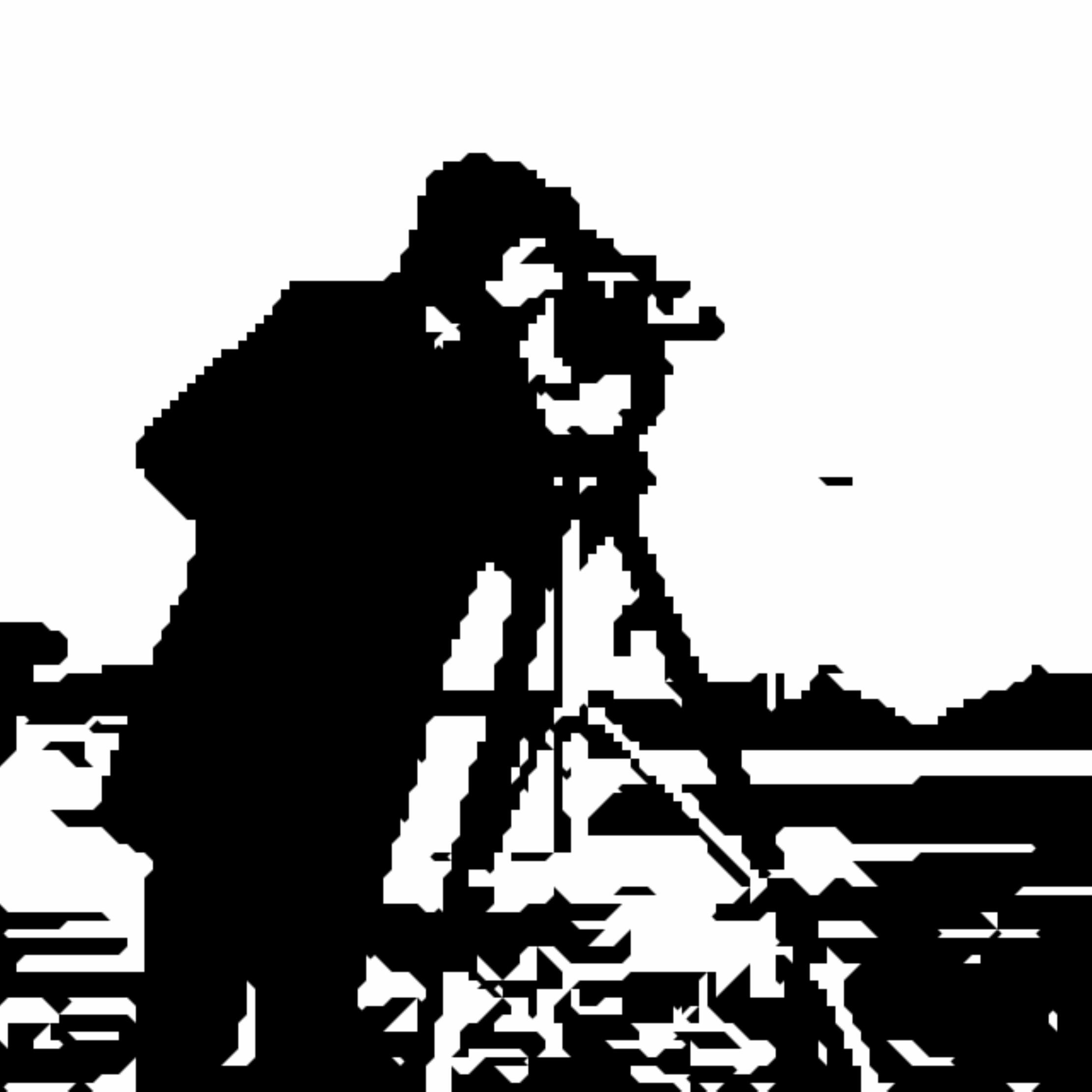} &
\includegraphics[width=4cm]{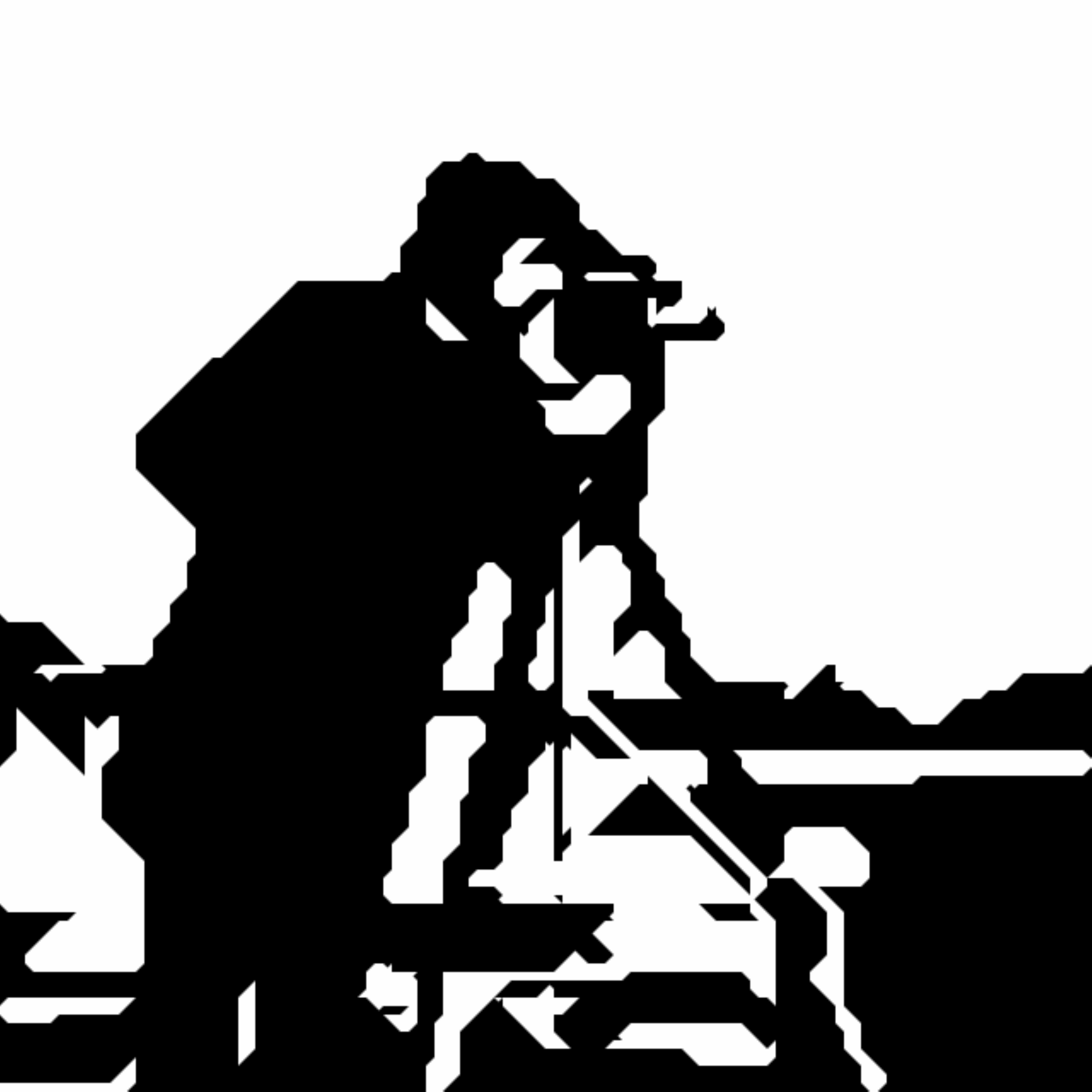} &
\includegraphics[width=4cm]{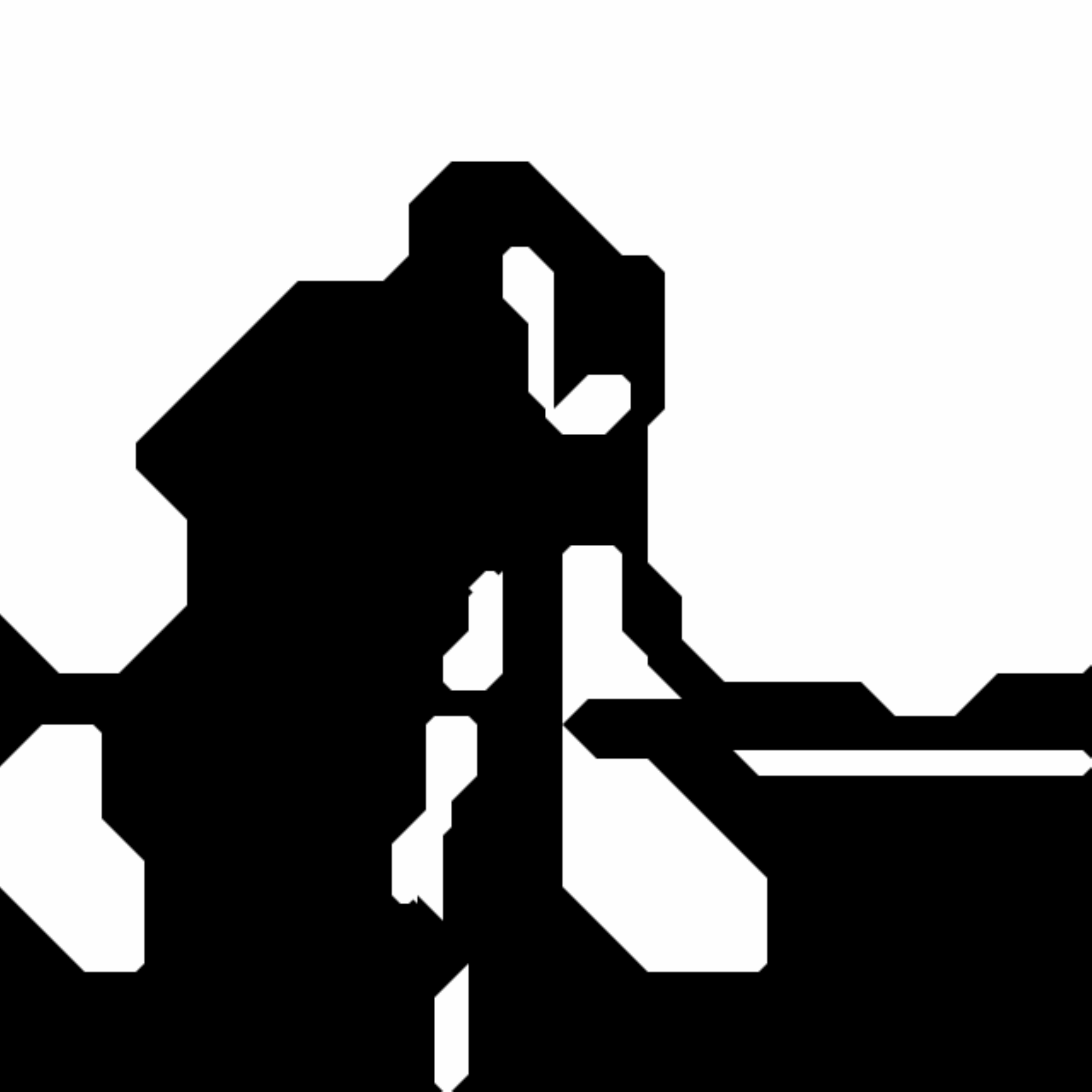} &
\\
{\small input} & {$\lambda = 1000, \nu = 10$} & {$\lambda = 10000, \nu = 10$} & {$\lambda = 100000, \nu = 10$}\\
& {\small within $0.75\%$.} & {\small within $3.6\%$.} & {\small within $4.4\%$.} 
\end{tabular}

\vspace{2mm}
\begin{tabular}{ccccc}
\includegraphics[width=4cm]{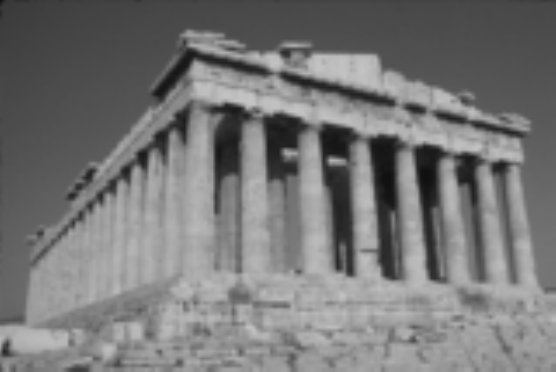} &
\includegraphics[width=4cm]{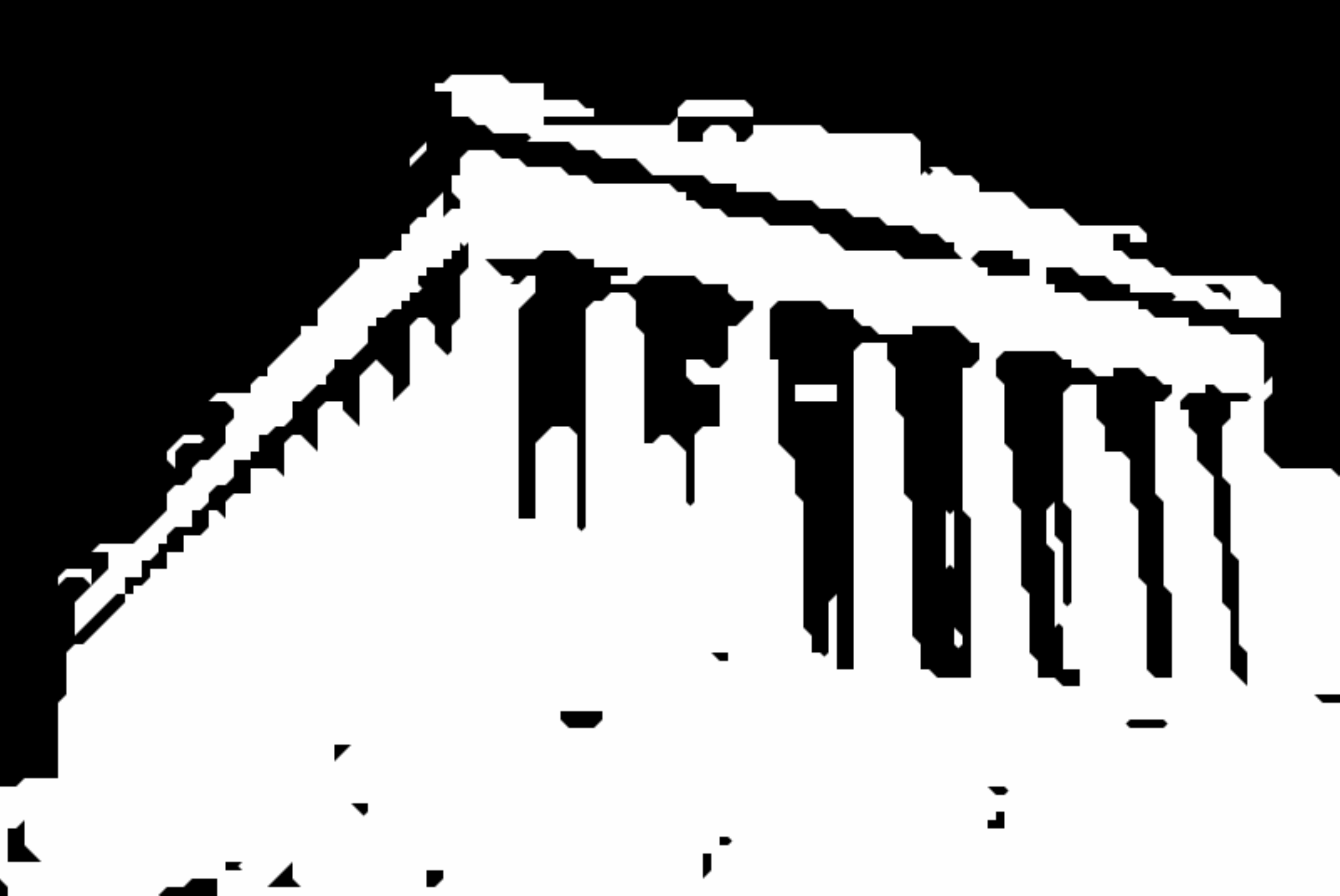} &
\includegraphics[width=4cm]{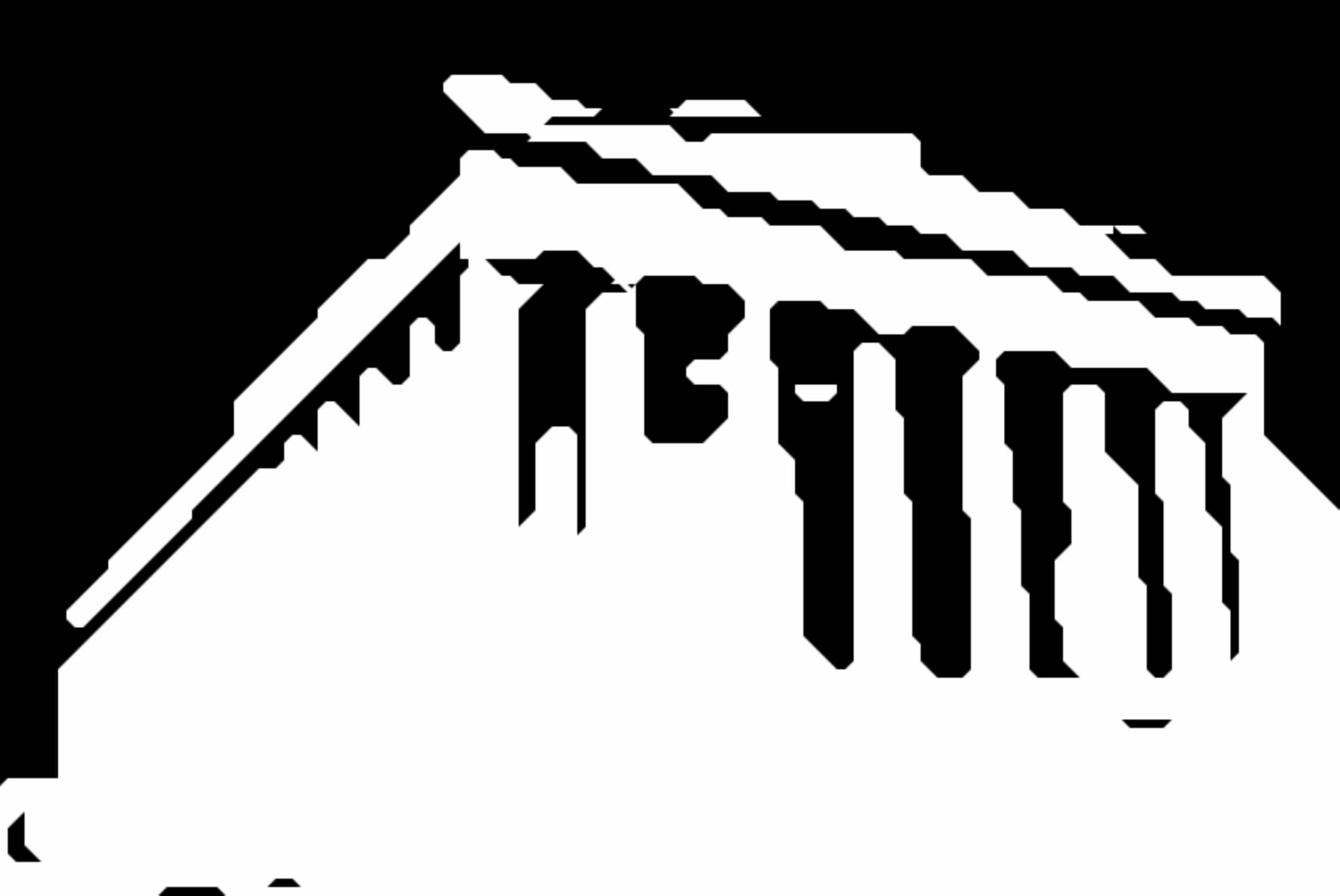} &
\includegraphics[width=4cm]{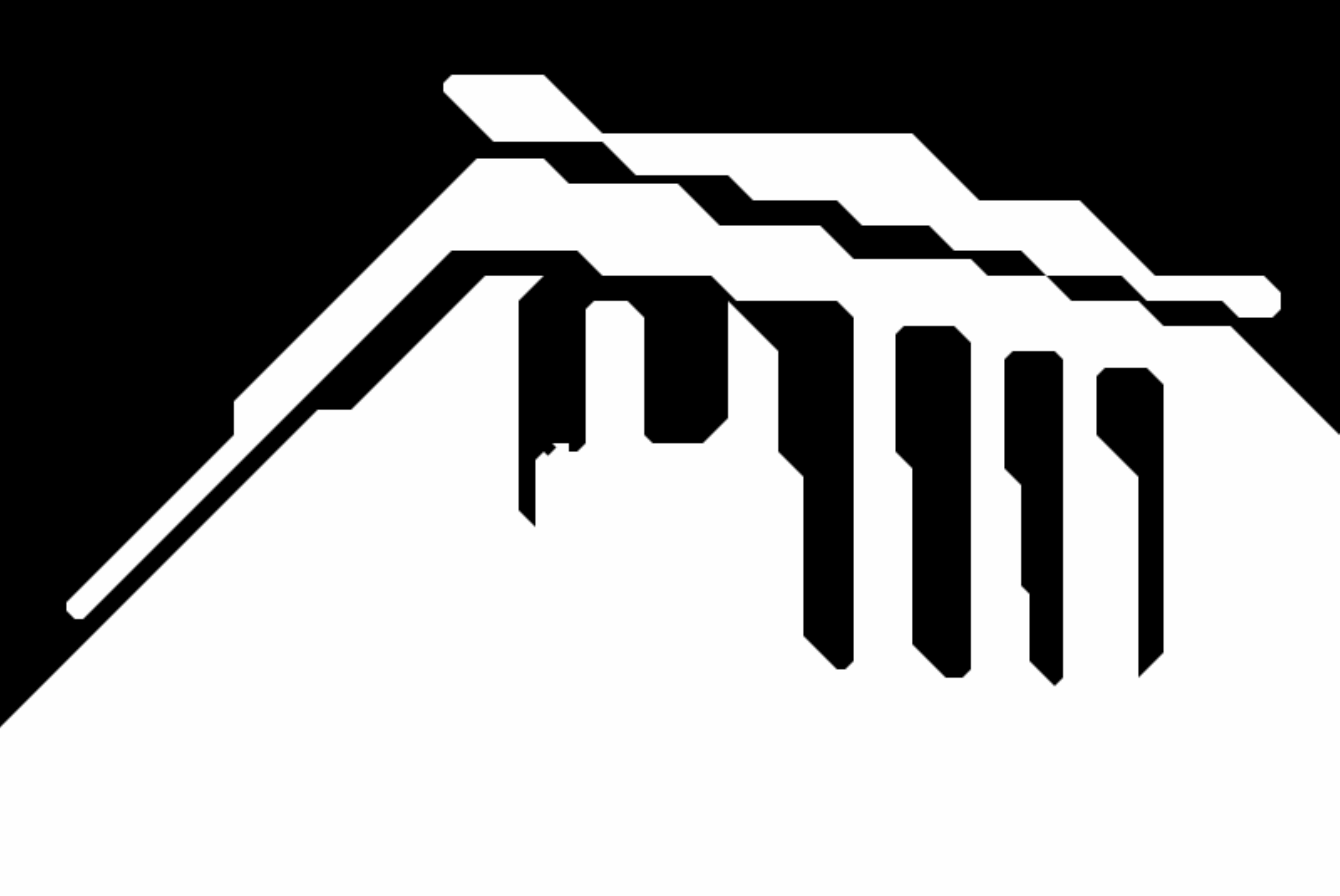}
\\
{\small input} & {$\lambda = 1000, \nu = 10$} & {$\lambda = 10000, \nu = 10$} & {$\lambda = 100000, \nu = 10$} \\
& {\small within $0.6\%$.} & {\small within $2.8\%$.} & {\small within $3.6\%$.}
\end{tabular}

\caption{Evaluation of the proposed method for unsupervised image segmentation and
an 8-connectivity. Shown are segmentations for different curvature weights and how close they
are to the lower bound (and hence the global optimum).}
\label{fig:curv_seg_exp}
\end{figure*}

Figure \ref{fig:curv_seg_exp} shows results of our method on images of
size $128 \times 128$ and $160 \times 107$, respectively, using an
8-connectivity. Here we provide results for different curvature
weights and it can be seen that even for very high weights long and
thin structures are preserved. At the same time, the relative gap
increases with the curvature weight.

These results took roughly $4$ hours computing time, where up to $9$
passes were needed. Though we re-used the existing solution, the first
passes often took as long as solving the initial program.

\paragraph{Interactive Image Segmentation}

Next we turn to the problem of interactive image segmentation for
color images, where in addition to an image $I : \Omega \rightarrow
\R^3$ we are given a set of foreground and background seed nodes as
specified by a user. Since our focus is on evaluating the novel
method, we did not refine the seed nodes. Hence, for each image the
seed nodes were specified only once.

From the given seed nodes, we estimate normalized histograms of color
values, resulting (after smoothing) in distributions $p_F(\cdot)$ and
$p_B(\cdot)$ that are then used in the model:
\begin{eqnarray}
&& - \int\limits_\Omega \! \log(p_F(I(\mf{x}))\, 
 [1 - u(\mf{x})]\, d\mf{x}\, \\
&&  -
\int\limits_\Omega\! \log(p_B(I(\mf{x}))\, u(\mf{x})\, d\mf{x}
\nonumber\\
&& \hspace{3mm} +\, \nu\, |C| + \lambda\! \int\limits_C  |\kappa_C(\mf{x})|^2\, d\mathcal{H}^1(\mf{x}) \ \ \ . \label{eq:interactive}
\end{eqnarray}
This function is minimized over all $u : \Omega \rightarrow [0,1]$ that
are consistent with the seed nodes.

For the experiments, we use a $16$-connectivity and fix the ratio of
the curvature weight $\lambda$ over the length weight $\nu$ to
$20.0$. The presented results were then obtained within half a day and
using between $6$ and $8$ GB of memory.

\begin{figure*}
\begin{center}
\begin{tabular}{cccc}
\includegraphics[width=3.8cm]{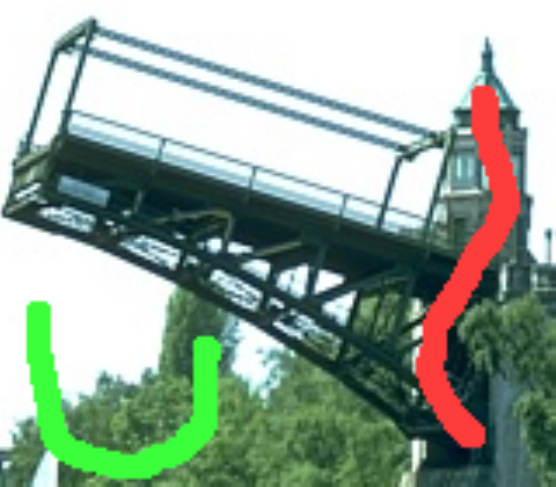} & 
\includegraphics[width=3.8cm]{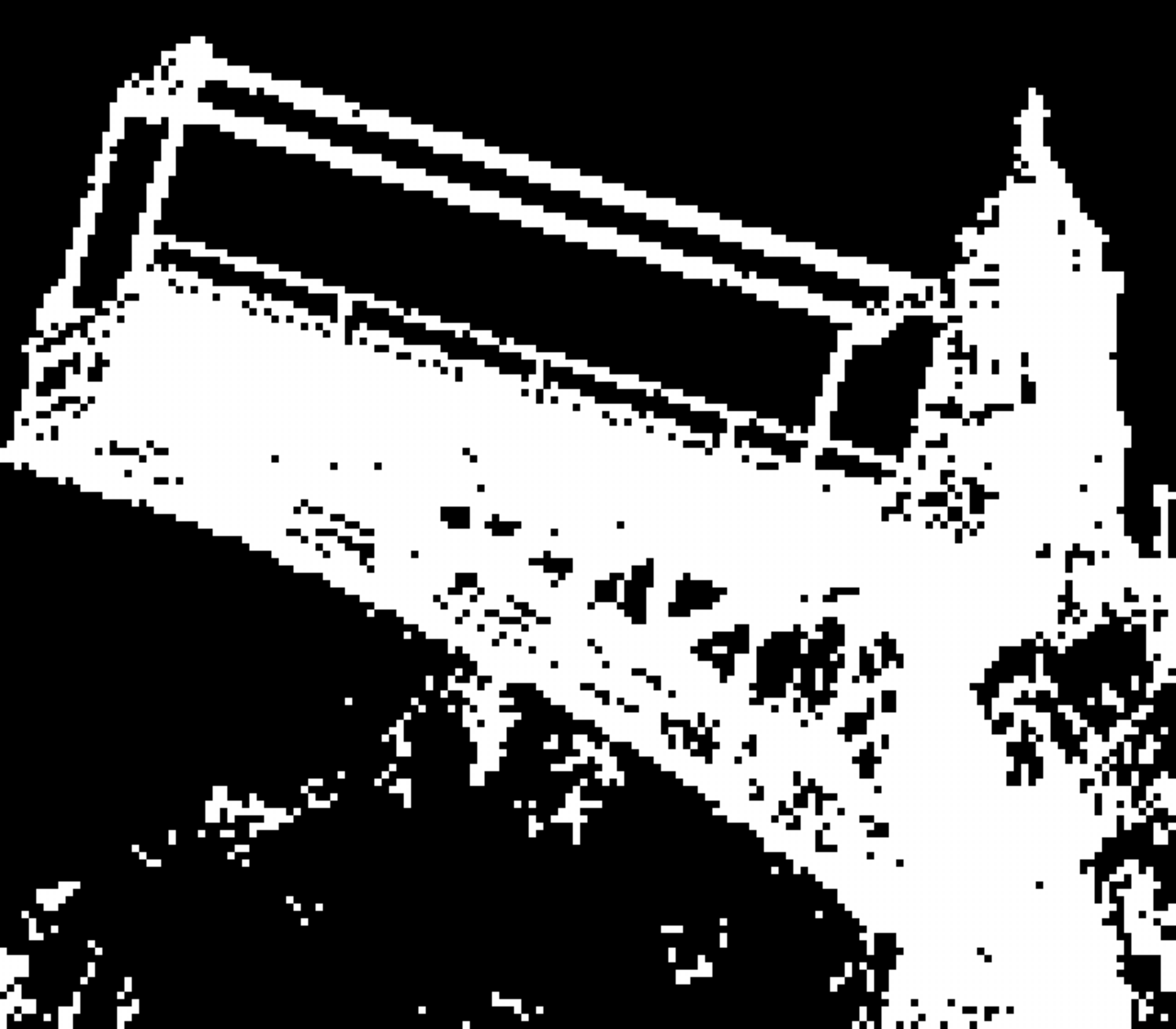} &
\includegraphics[width=3.8cm]{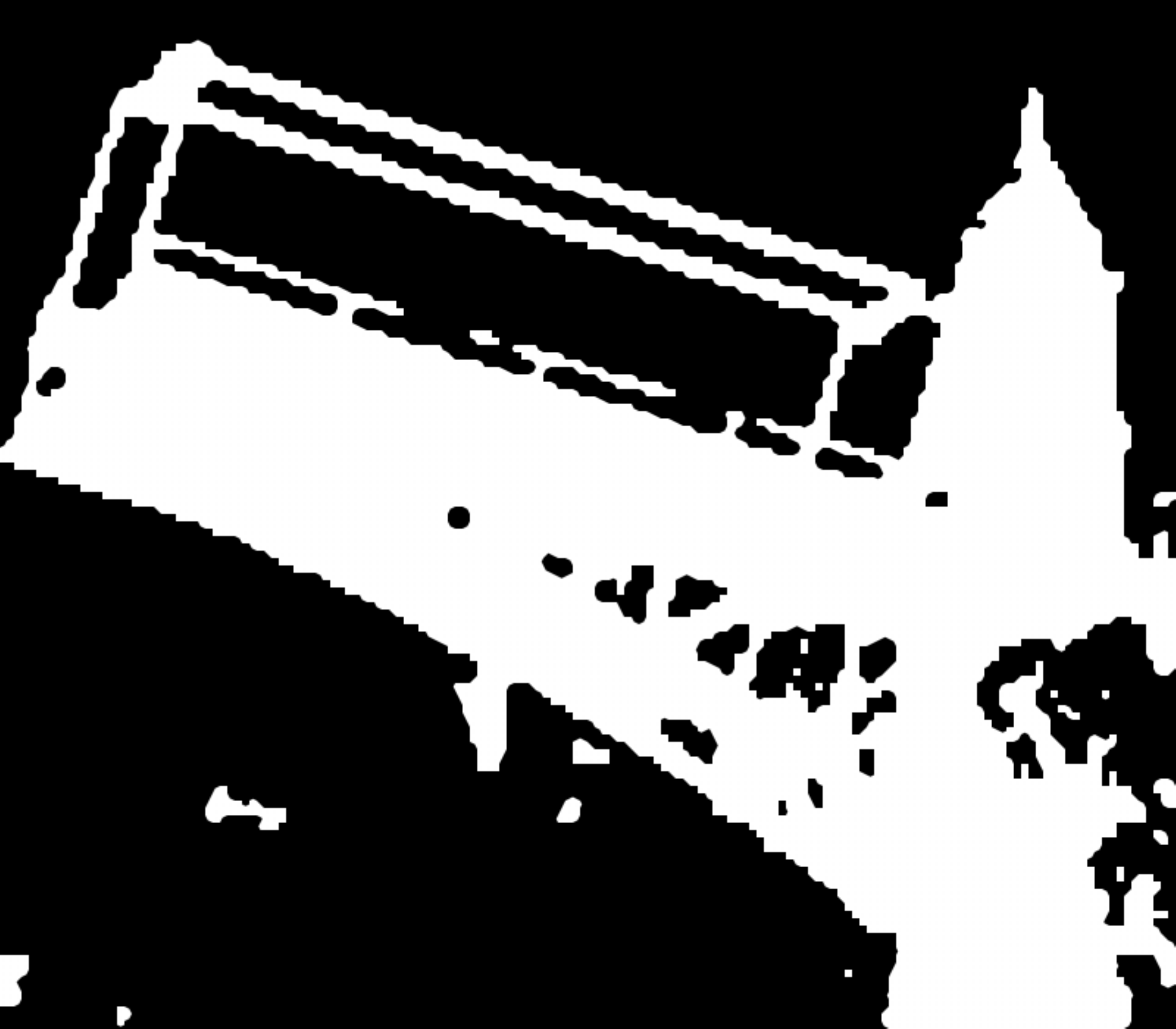} &
\includegraphics[width=3.8cm]{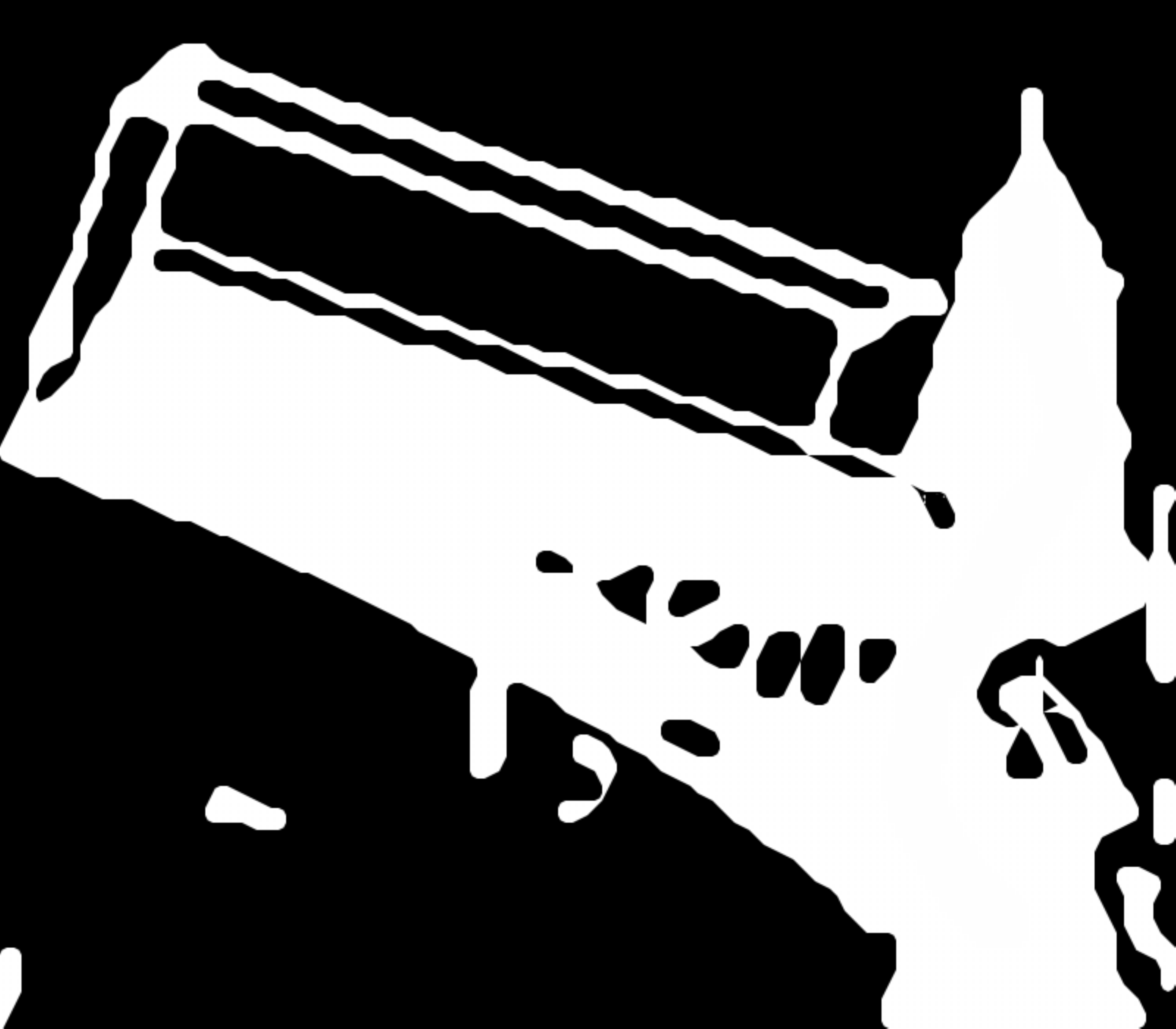}\\[-.5mm]
&&& {\small within $0.5\%$.}\\[1mm]
\includegraphics[width=3.8cm]{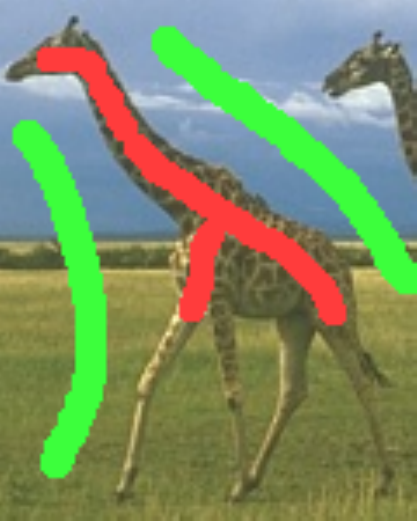} &
\includegraphics[width=3.8cm]{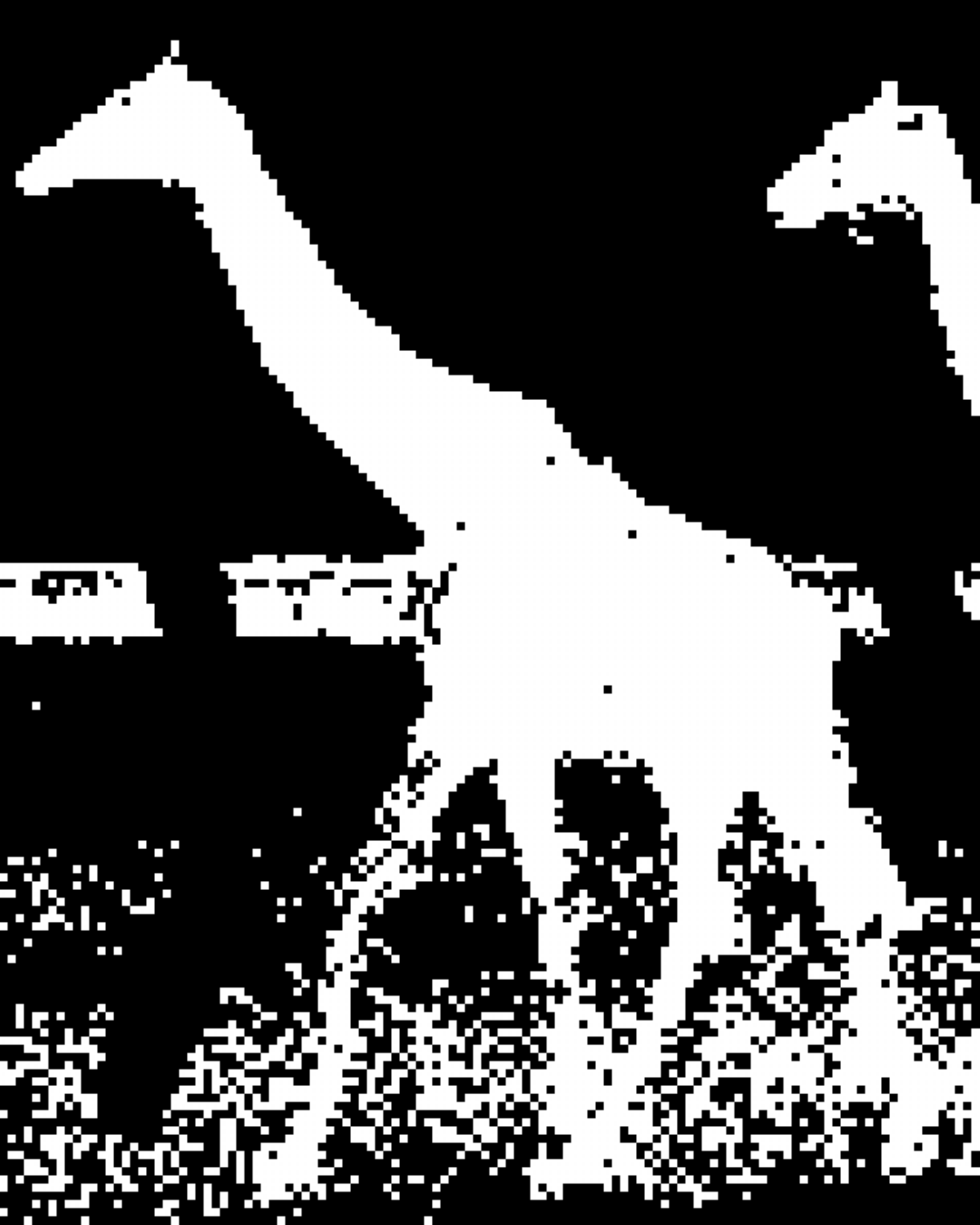} &
\includegraphics[width=3.8cm]{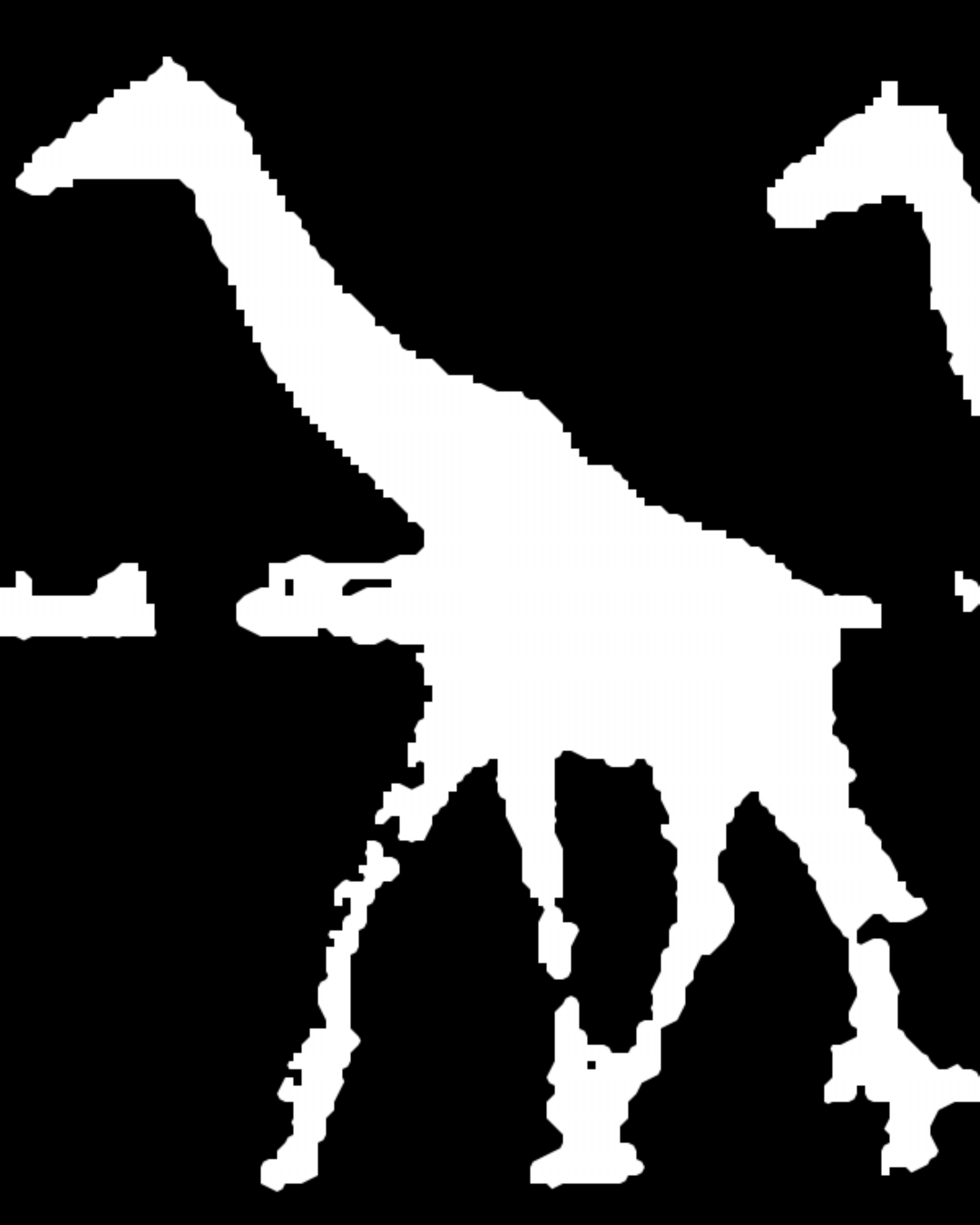} &
\includegraphics[width=3.8cm]{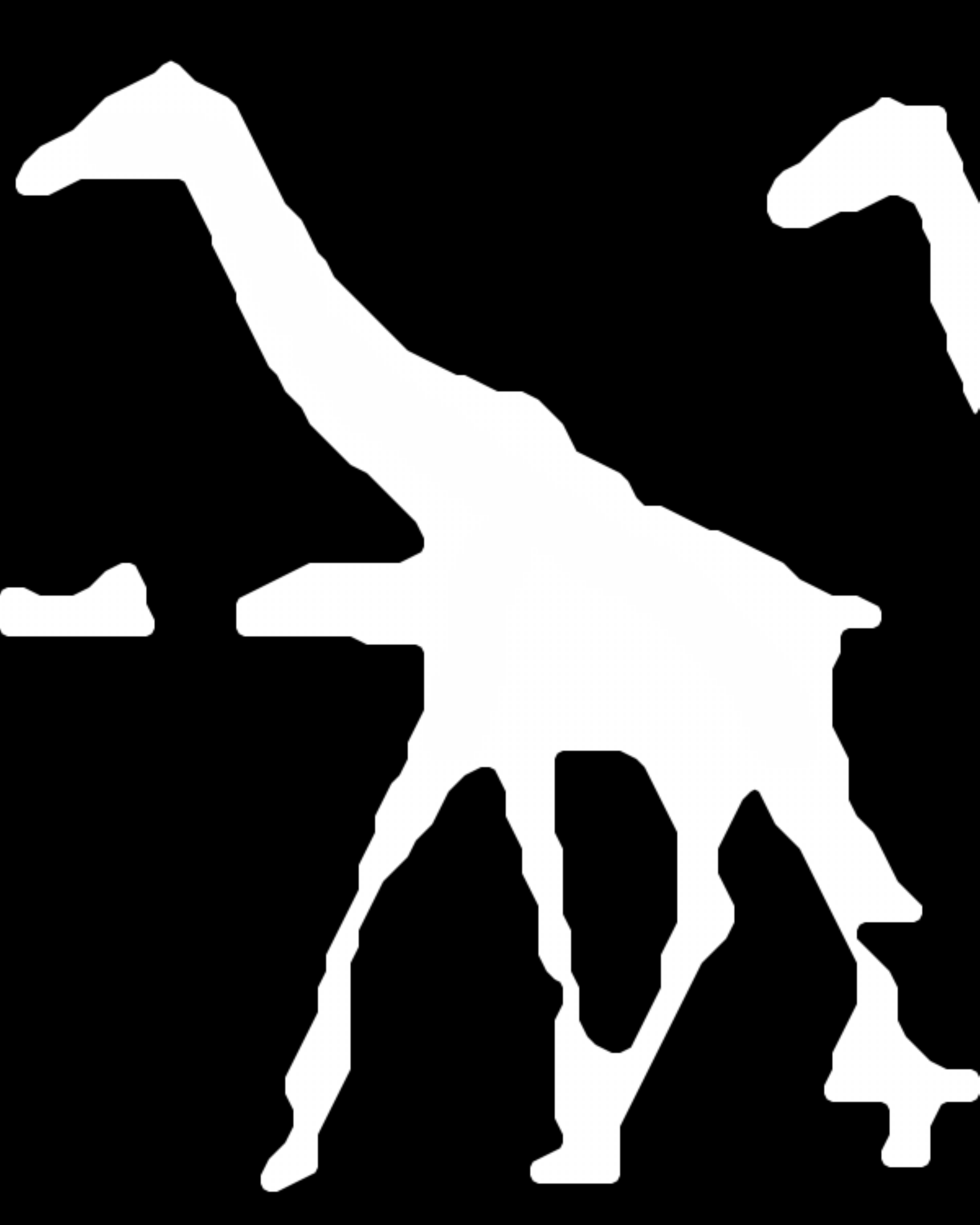}
\\[-.5mm]
&&& {\small \textbf{global optimum}}\\[1mm]
\includegraphics[width=3.8cm]{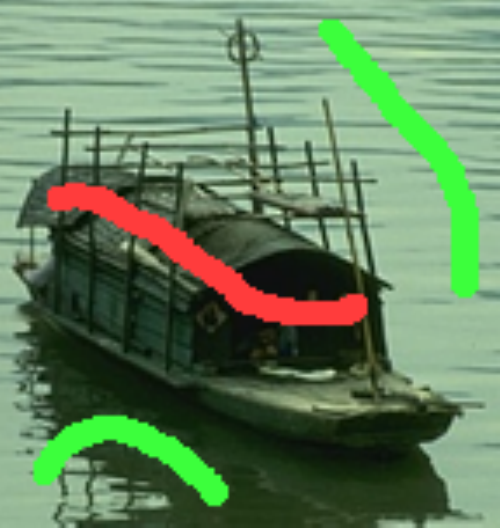} &
\includegraphics[width=3.8cm]{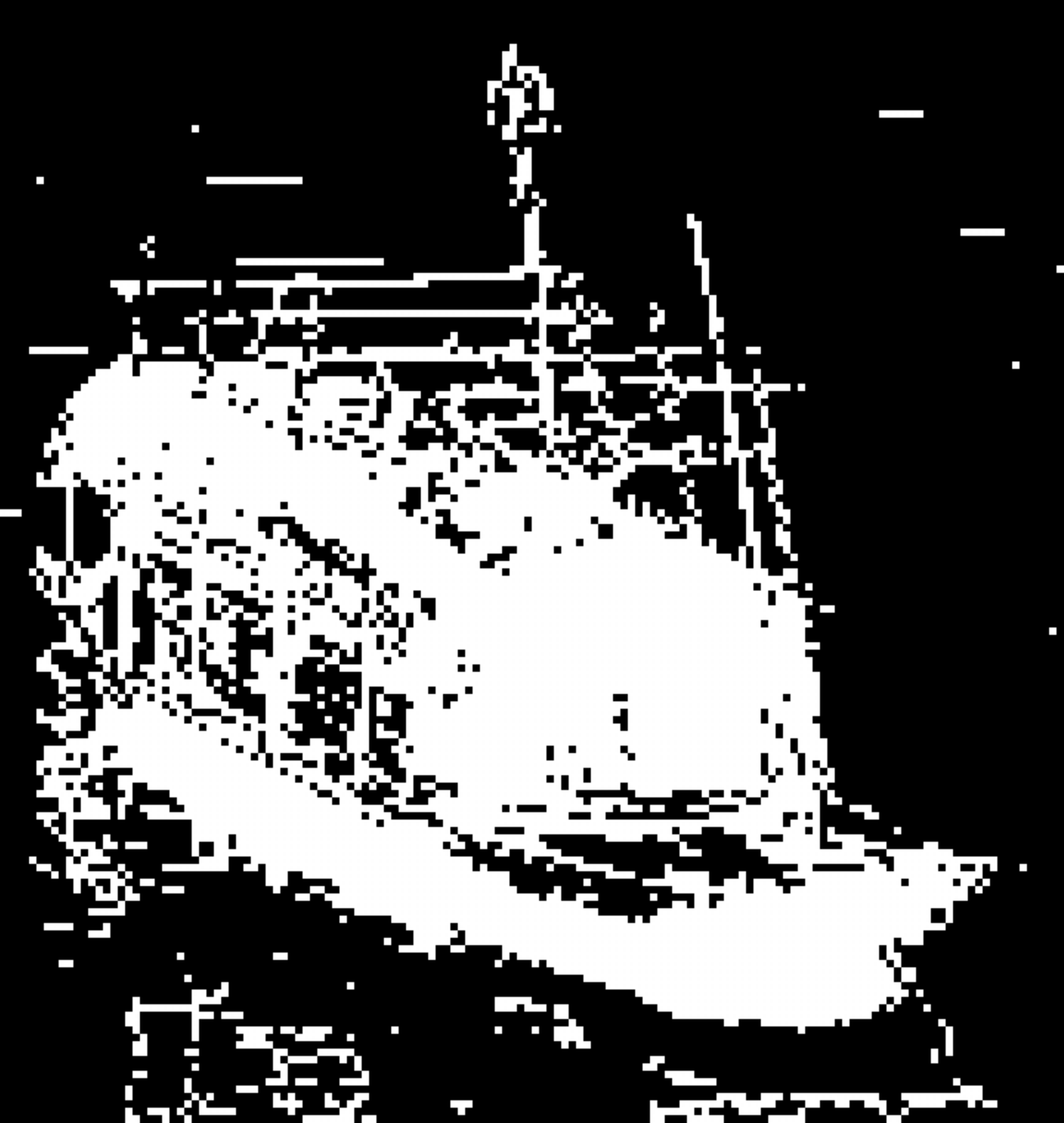} &
\includegraphics[width=3.8cm]{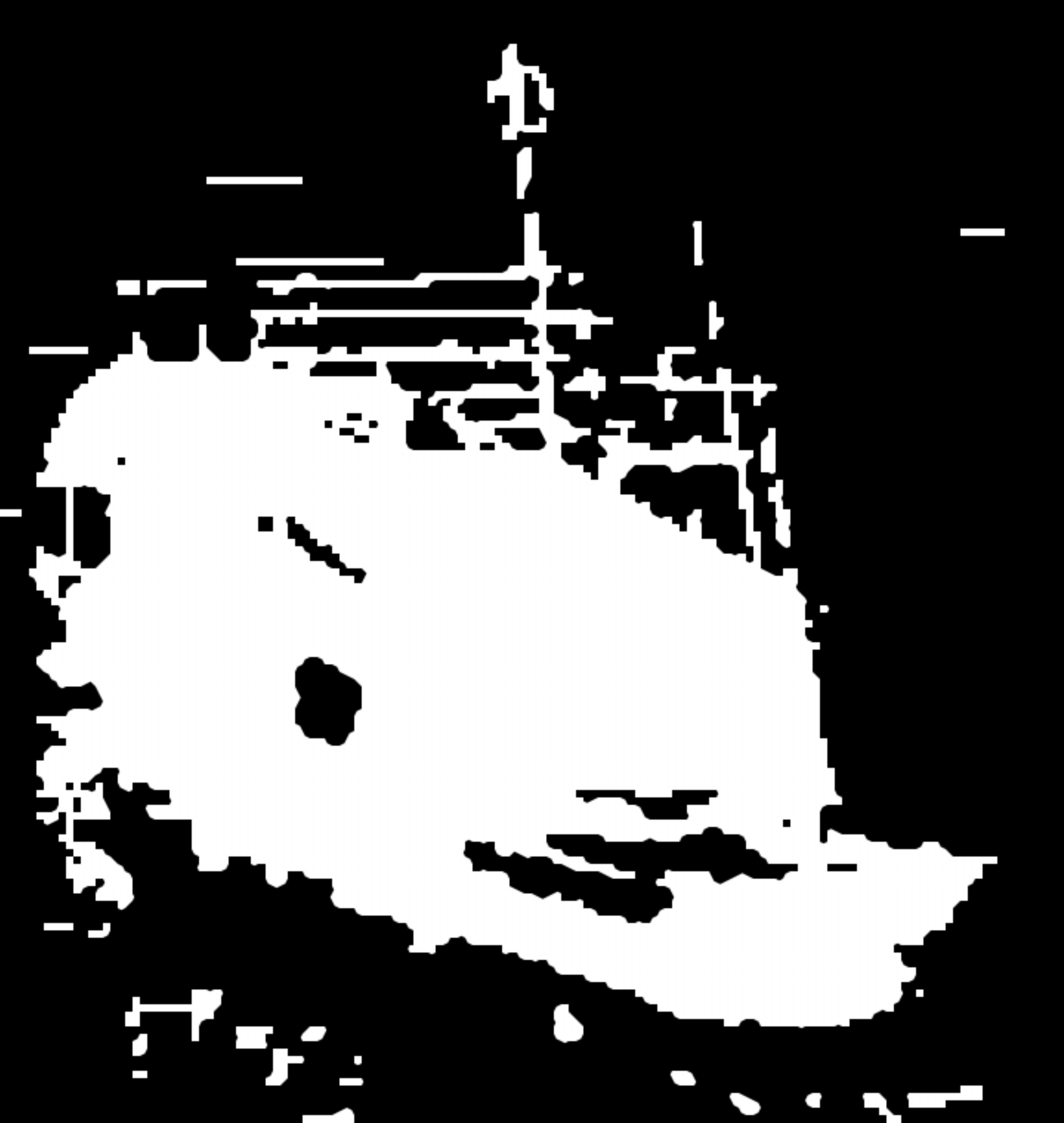} &
\includegraphics[width=3.8cm]{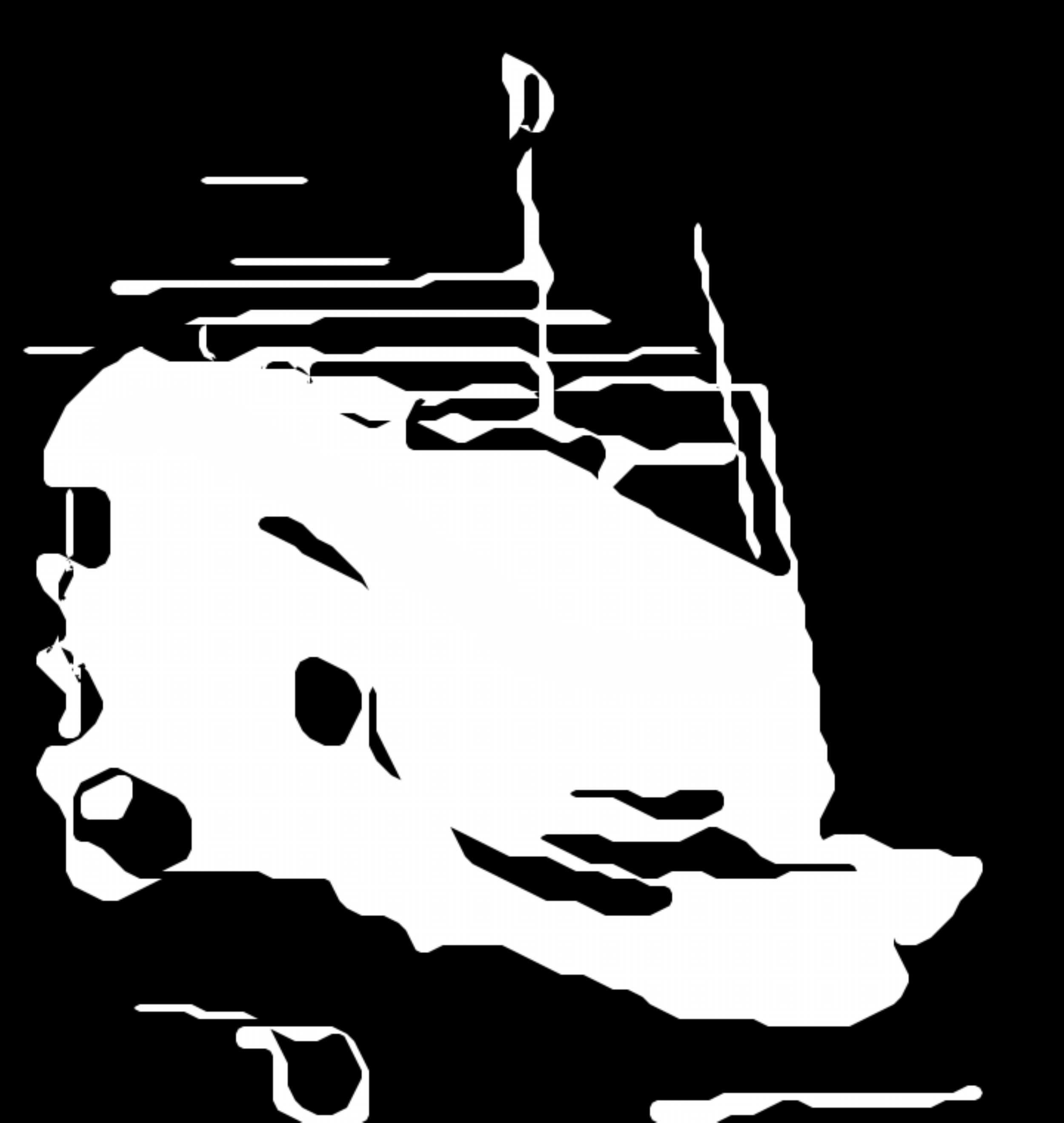} \\[-0.5mm]
&&& {\small within $2.6\%$.}\\[1mm]
\includegraphics[width=3.8cm]{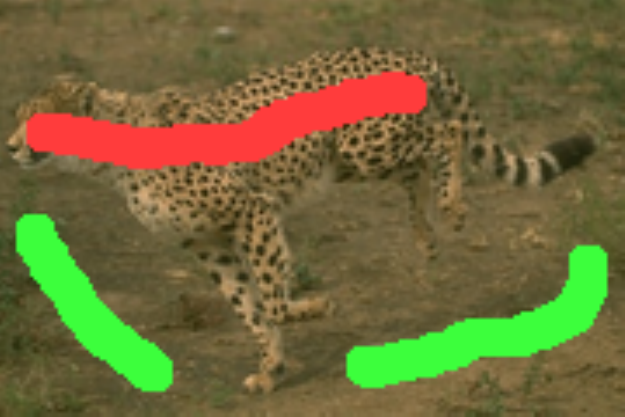} &
\includegraphics[width=3.8cm]{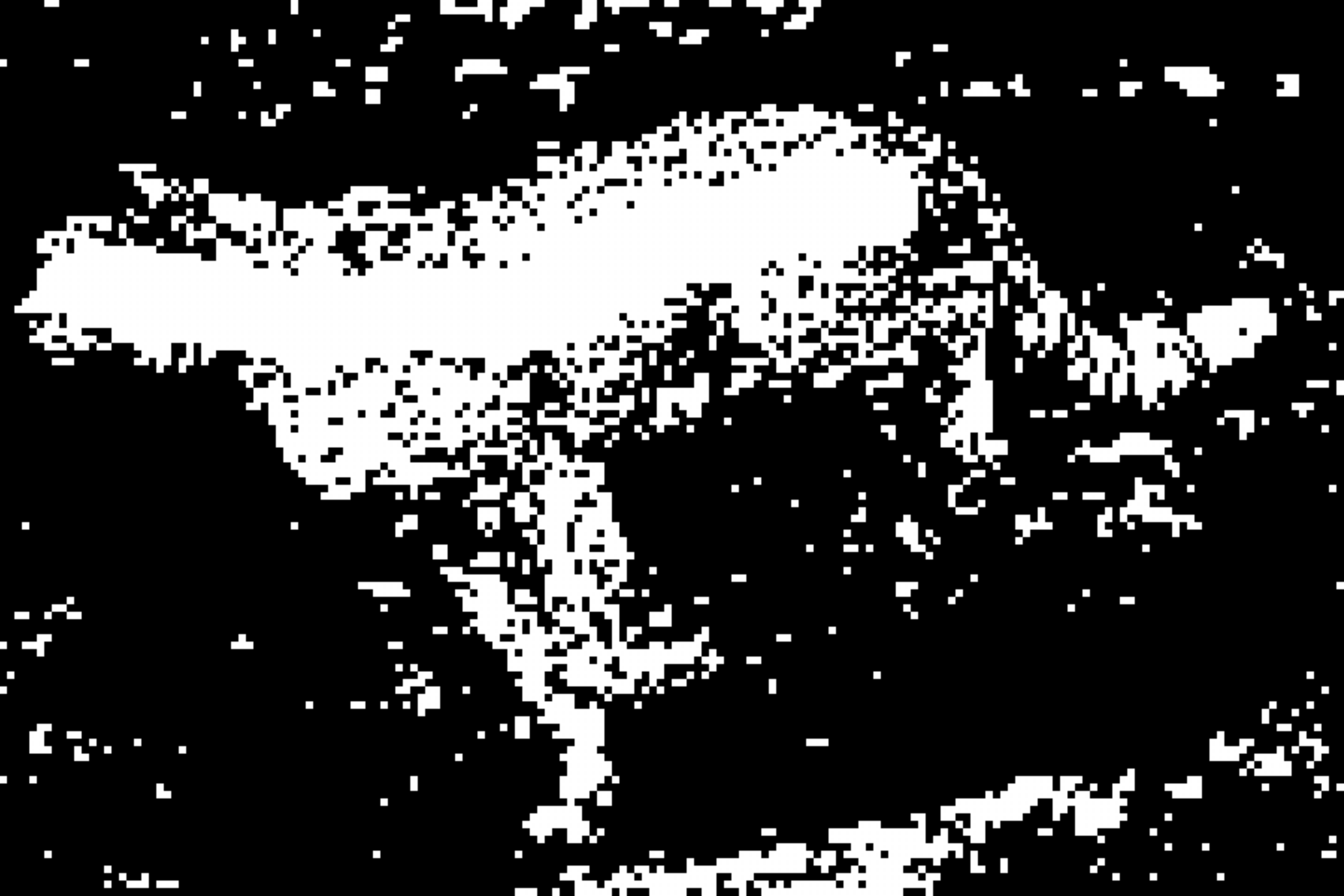} &
\includegraphics[width=3.8cm]{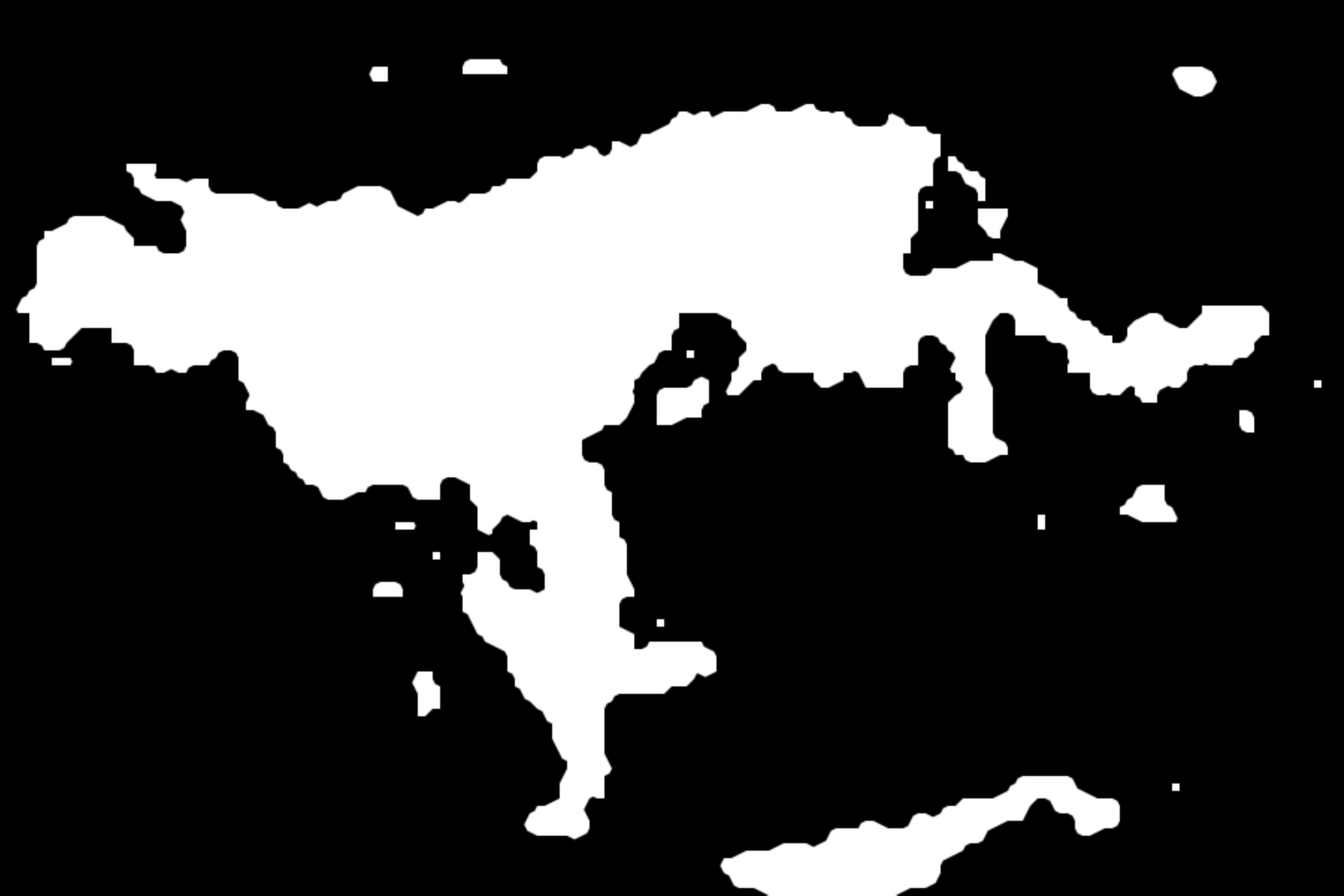} &
\includegraphics[width=3.8cm]{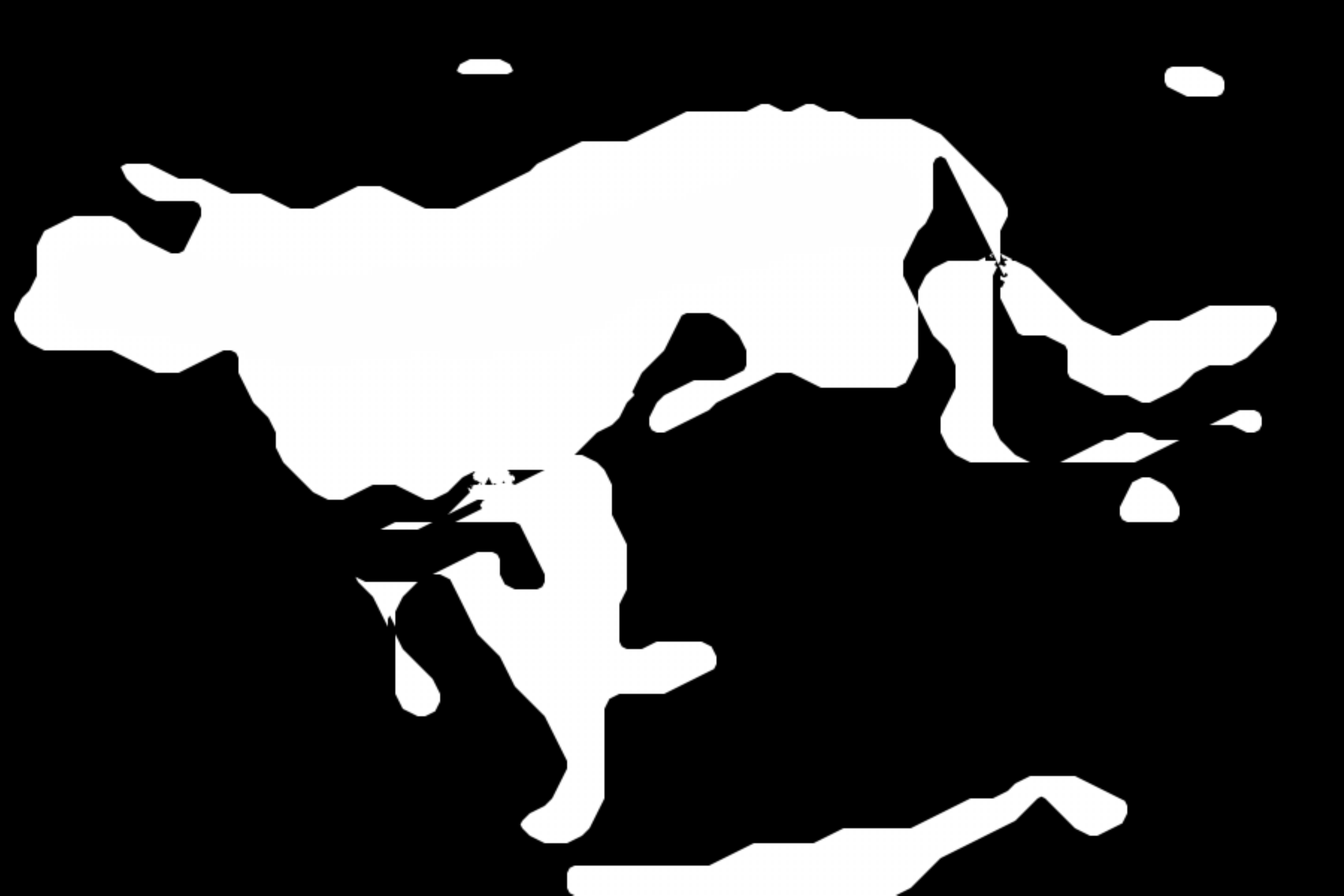} \\[-0.5mm]
&&& {\small within $2.2\%$.}\\[1mm]
\includegraphics[width=3.8cm]{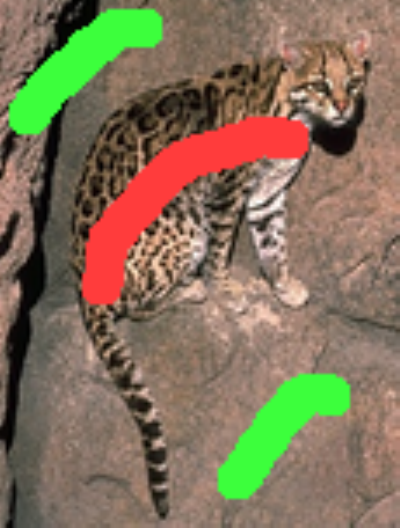} &
\includegraphics[width=3.8cm]{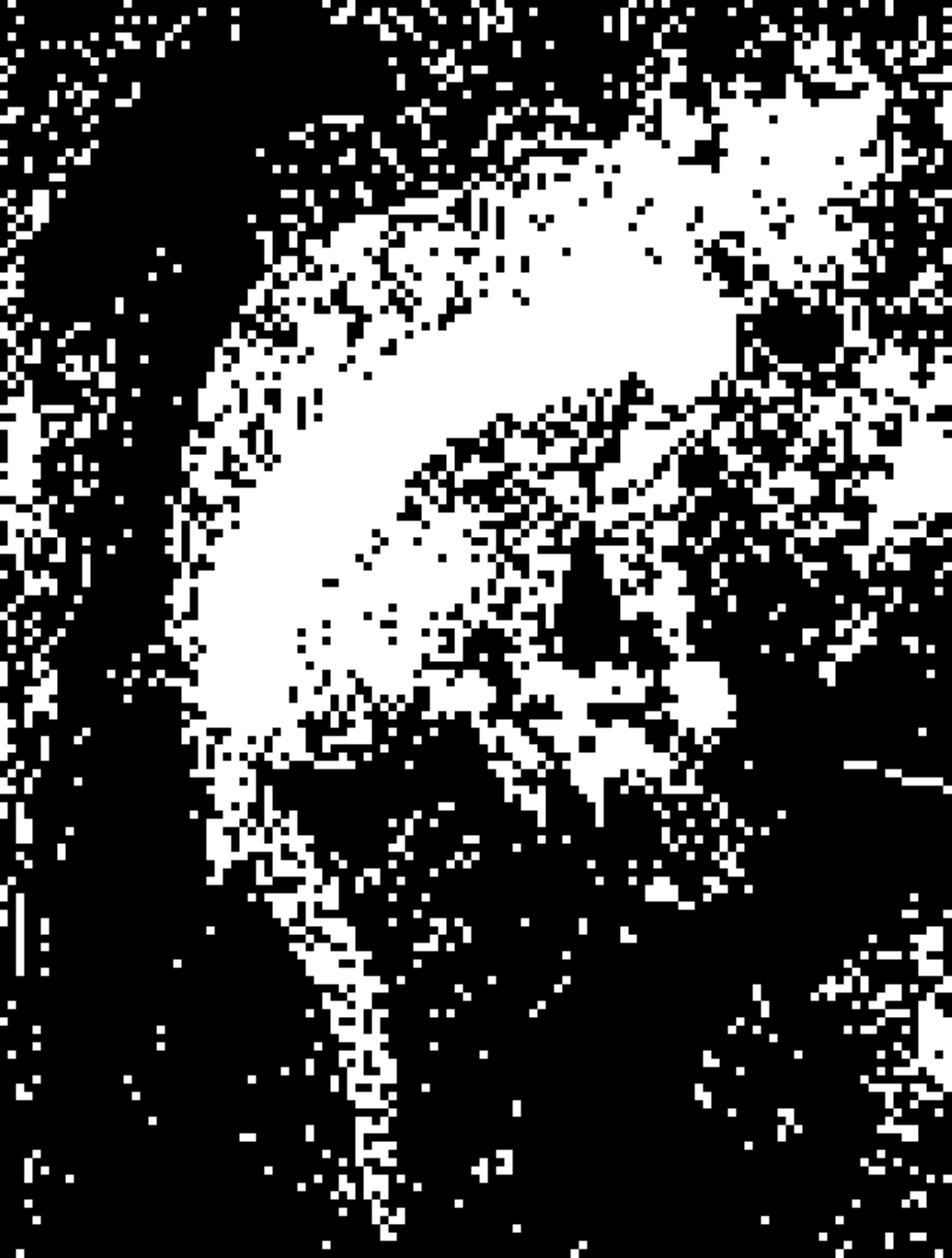} &
\includegraphics[width=3.8cm]{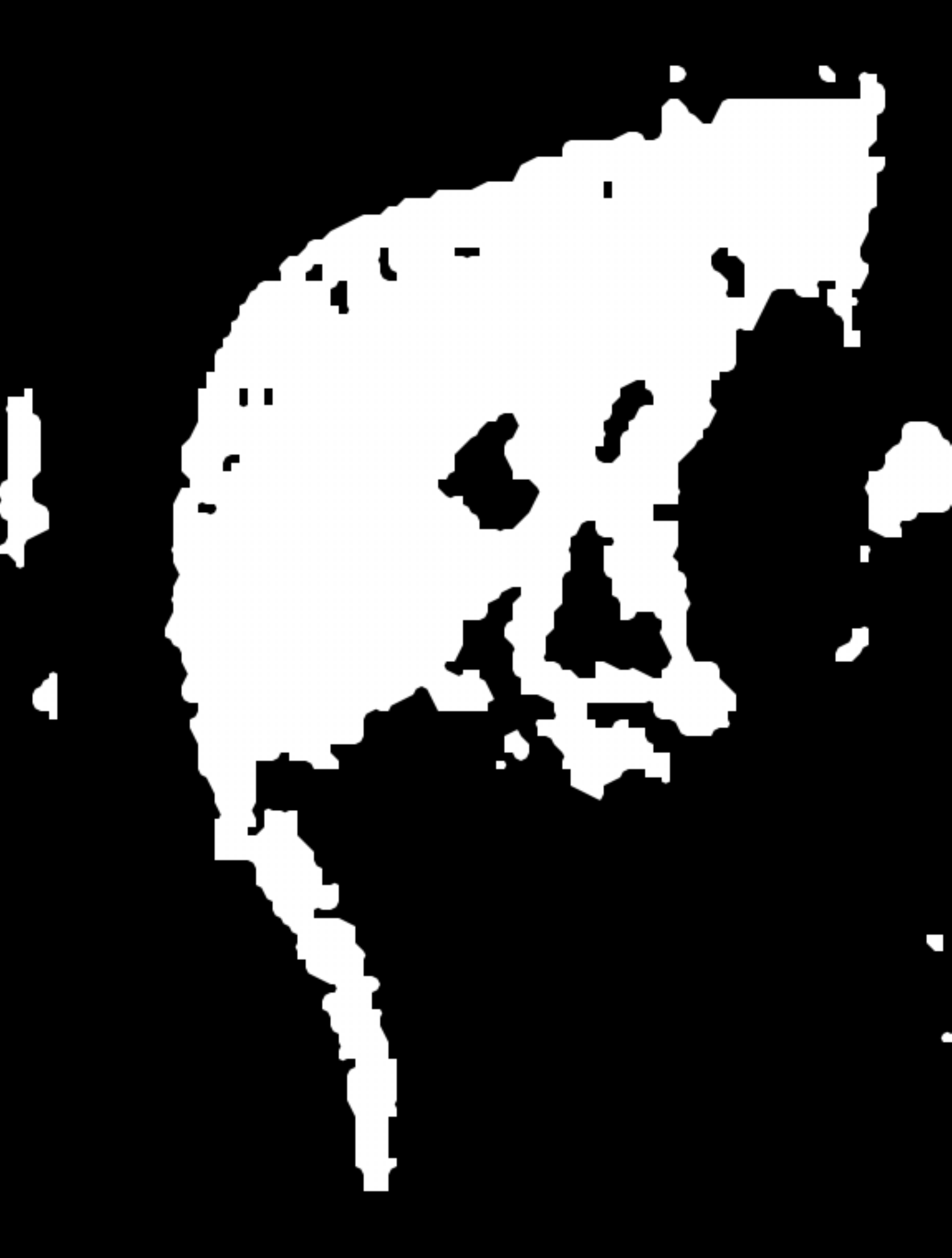} &
\includegraphics[width=3.8cm]{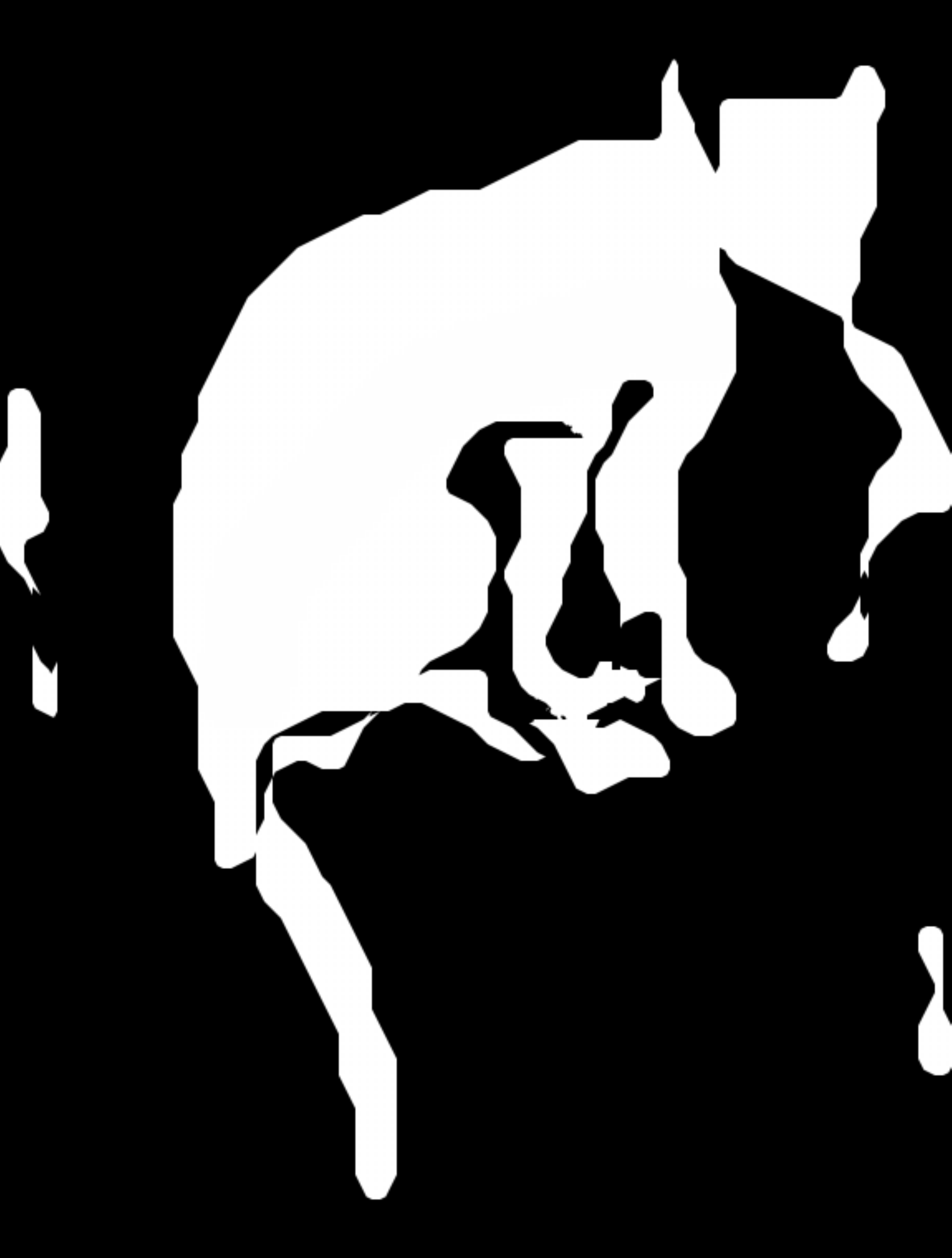} \\[-0.5mm]
&&& {\small within $1.5\%$.}\\[1mm]
\end{tabular}
\caption{Comparison of length-based and curvature-based methods. Left column: input images with seed nodes super-imposed.
Middle left: thresholding scheme. Middle right: length-based
segmentation (proposed method). In all cases $\lambda = 1$. Right:
length- and curvature-based segmentation (proposed method) and how
close the results are to the global optimum. In all cases we set
$\lambda = 4$ and $\nu = 0.2$.  Images taken from the Berkeley
database
\texttt{http://www.eecs.berkeley.edu/Research/Projects/CS/vision/grouping/segbench/}.}
\label{fig:interactive}
\end{center}
\end{figure*}

Figure \ref{fig:interactive} compares our results with a simple
thresholding scheme and the length based method of
Section~\ref{sec:len}.  These results clearly show the tendency of
length-based methods to suppress long and thin objects. With curvature
regularity this is remedied.

\paragraph{Effects of the Constraint Sets}

\begin{figure}
\begin{center}
\setlength{\tabcolsep}{1mm}
\begin{tabular}{cc}
\includegraphics[width=0.49\linewidth]{boat_crop_n16_gamma4_lambda0_2_edgecon_crosscon.pdf} &
\includegraphics[width=0.49\linewidth]{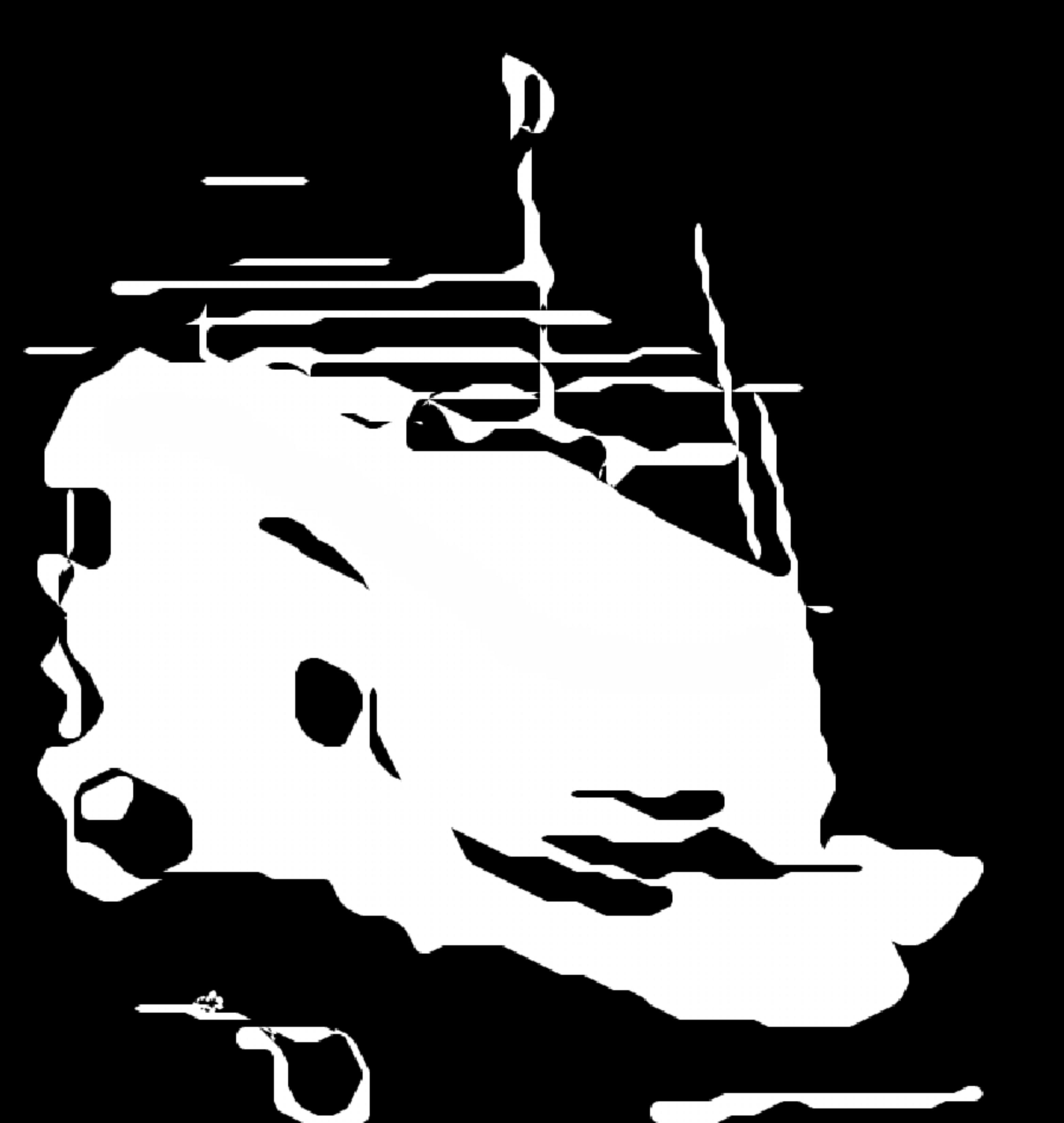}\\[-1mm]
{\small with crossing prevention} &  {\small without crossing prevention}\\[1mm]
\includegraphics[width=0.49\linewidth]{leo1_n16_gamma4_lambda0_2_edgecon_crosscon.pdf} &
\includegraphics[width=0.49\linewidth]{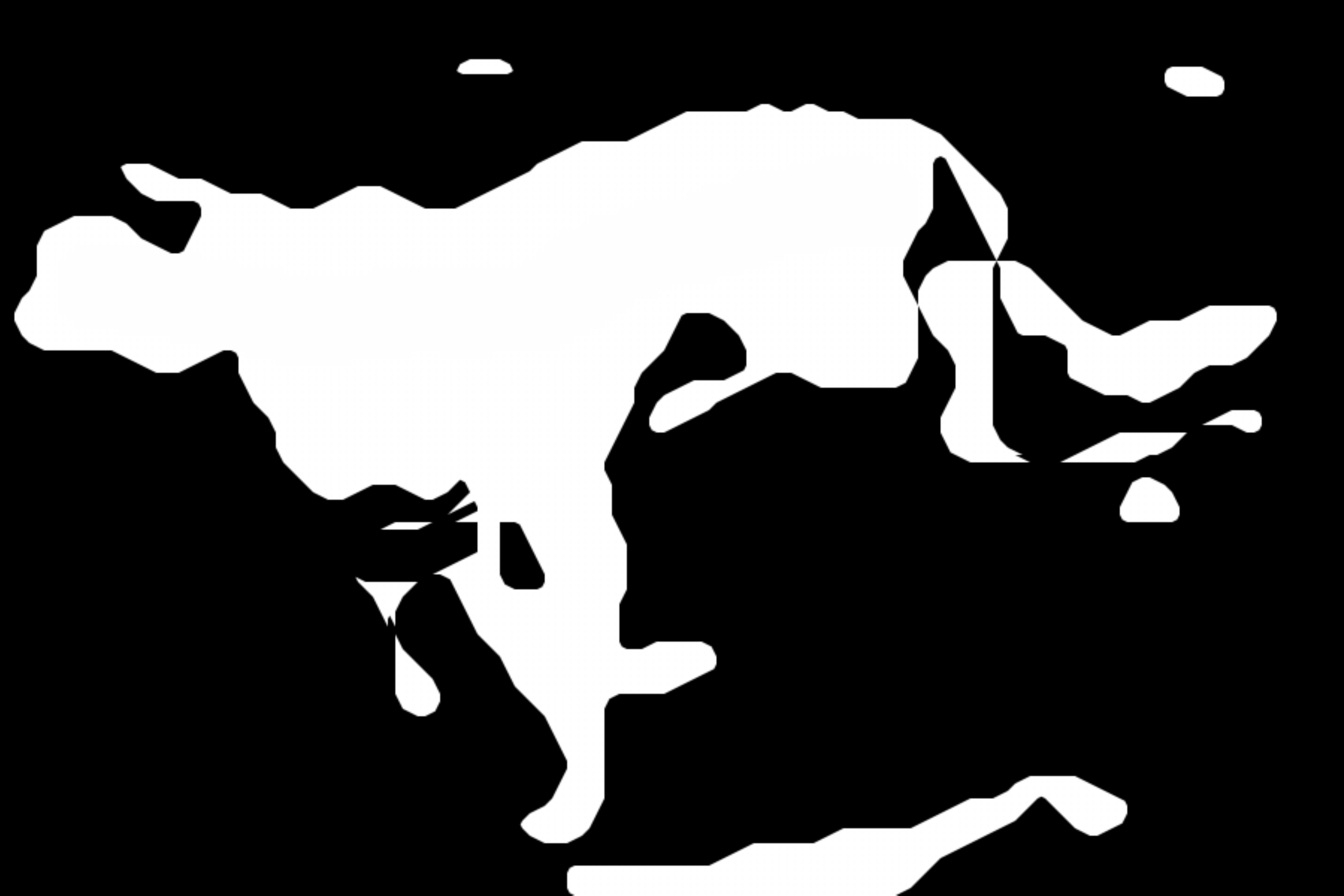}
\\[-1mm]
{\small with crossing prevention} &  {\small without crossing prevention}
\end{tabular}
\caption{Without crossing prevention there is a bias towards triple points. For the top row 
significantly different segmentations are obtained with and without
the constraints. For the bottom row there are only minor changes. }
\label{fig:cross_prev}
\end{center}
\end{figure}

Above we indicated that it is debatable whether self-intersecting
region boundaries should be allowed or not. In Figure
\ref{fig:cross_prev} both possibilities are explored. For these images
there were significant differences, for others, like the giraffe
image, none at all. For the boat the results clearly improve when
forbidding self-intersections, for the other image they grow slightly
worse. In future work we hope to get closer to the respective global
optima to find out which model is actually better.

Lastly, in this work we added the boundary consistency constraints to
the original formulation in \cite{Schoenemann-Kahl-Cremers-09}. The
latter formulation did in fact not represent the model
correctly. Still, due to the relaxation adding the new constraints
does not entirely solve the problem. When comparing the thresholded
solutions of both versions we noticed as many changes to the good as
to the bad. In future work we hope to improve on this.

\subsection{Inpainting}

\begin{figure*}
\begin{center}
\begin{tabular}{ccc}
\includegraphics[width=0.3\textwidth]{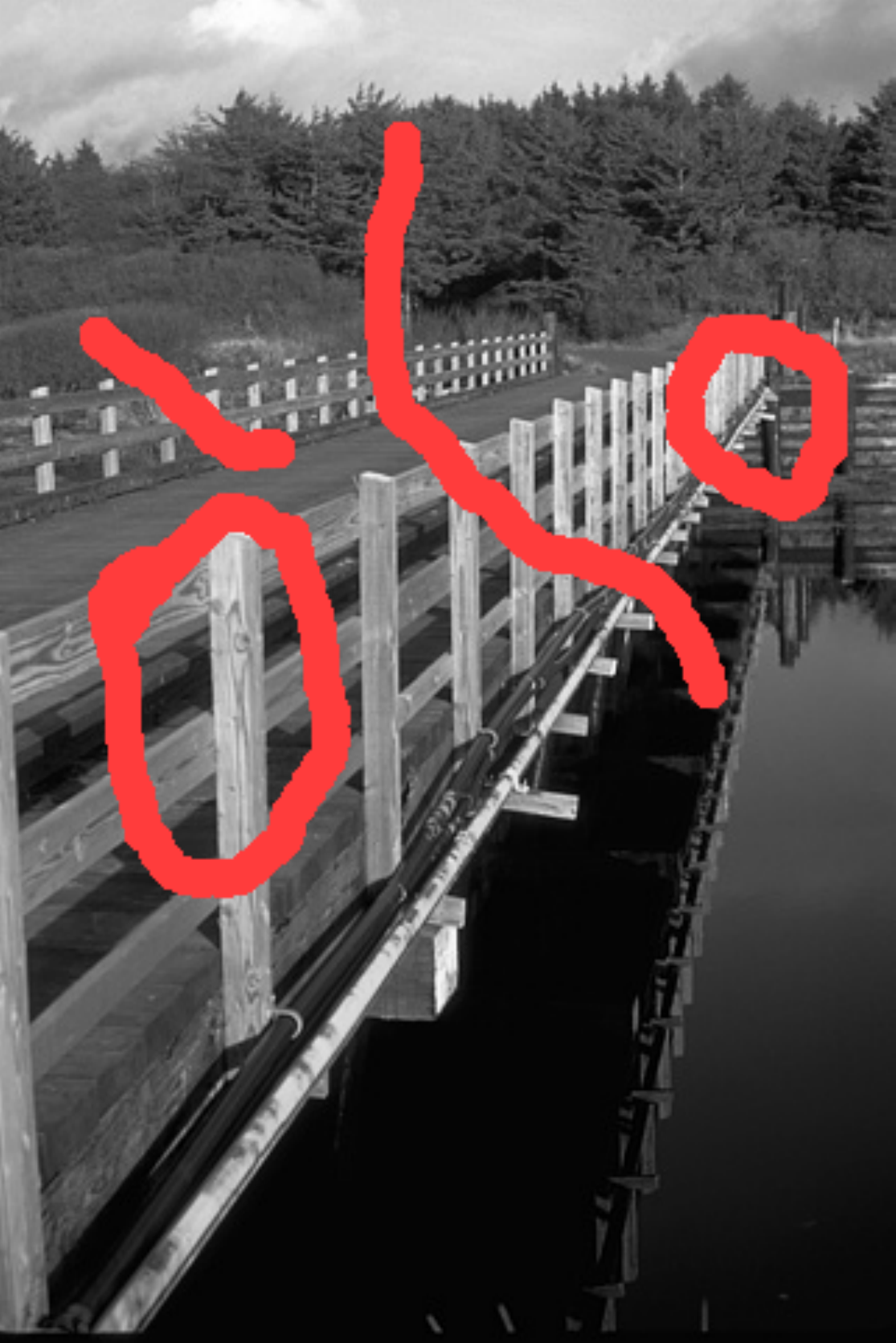} &
\includegraphics[width=0.3\textwidth]{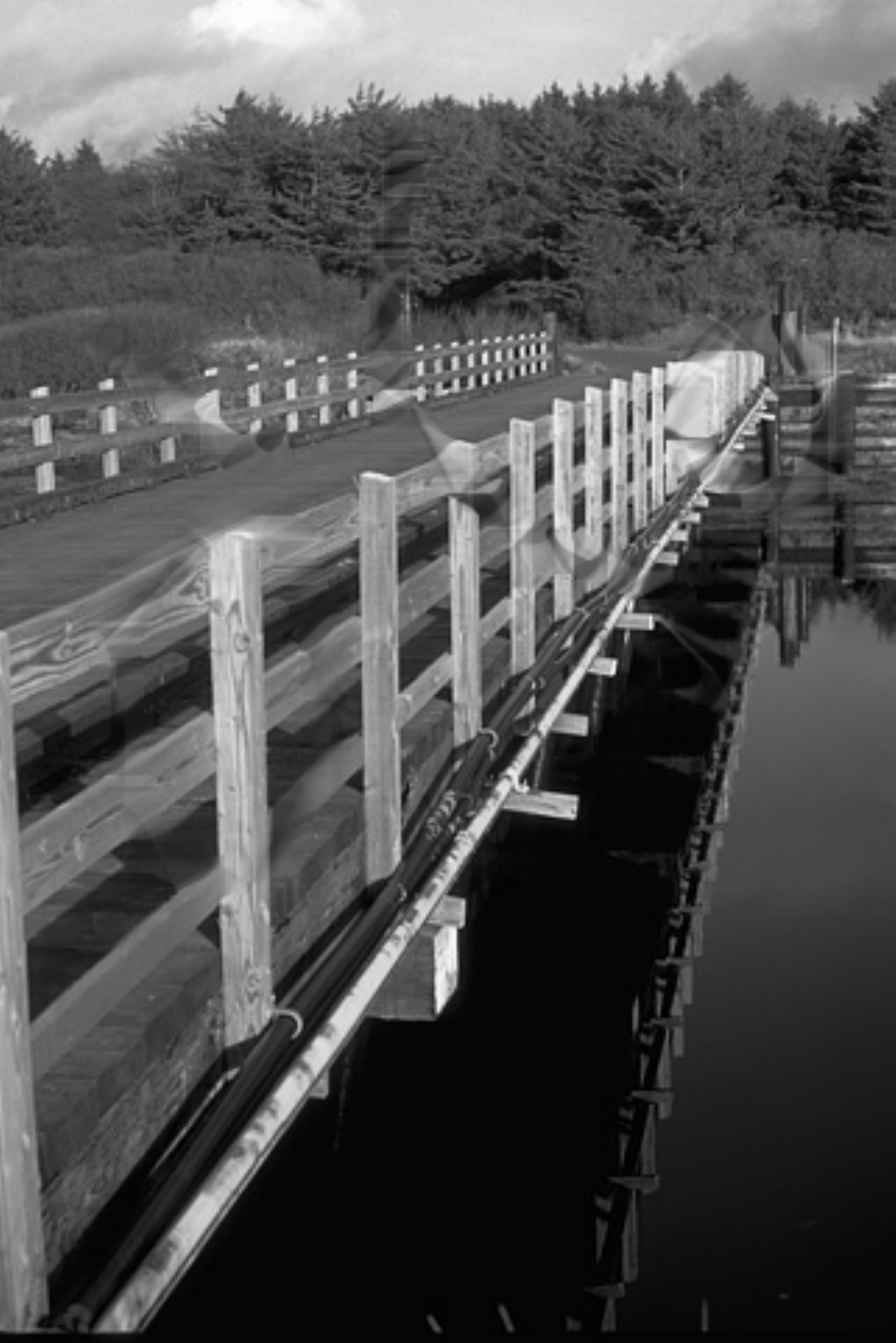} &
\includegraphics[width=0.3\textwidth]{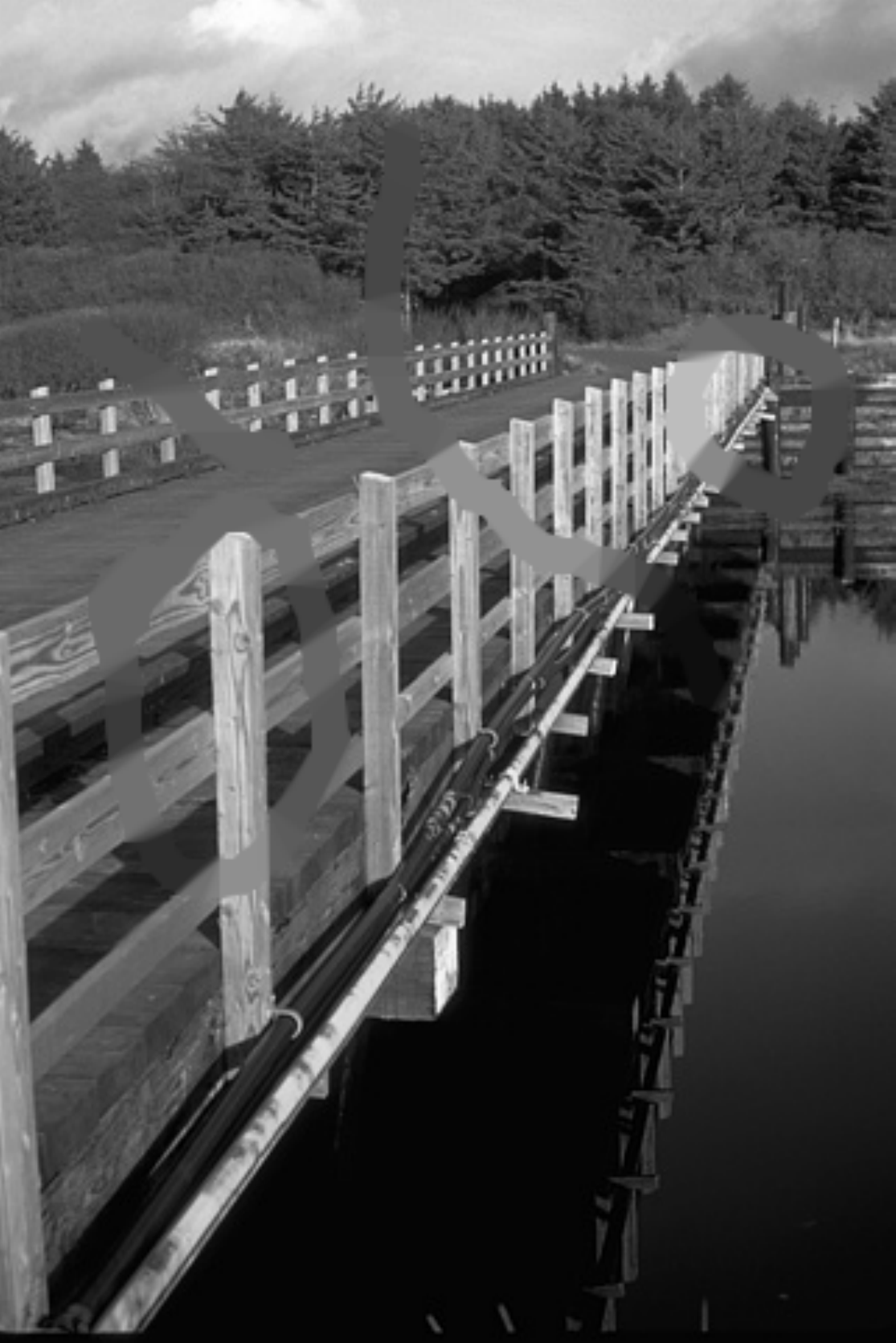}\\[-0.5mm]
{\small damaged image} & {\small inpainted with squared curvature} & 
{\small inpainted with length}
\end{tabular}
\caption{For inpainting curvature is much better suited than length regularity.}
\label{fig:bridge_inpainting}
\end{center}
\end{figure*}

\begin{figure*}
\begin{center}
\begin{tabular}{cc}
\includegraphics[width=0.4\textwidth]{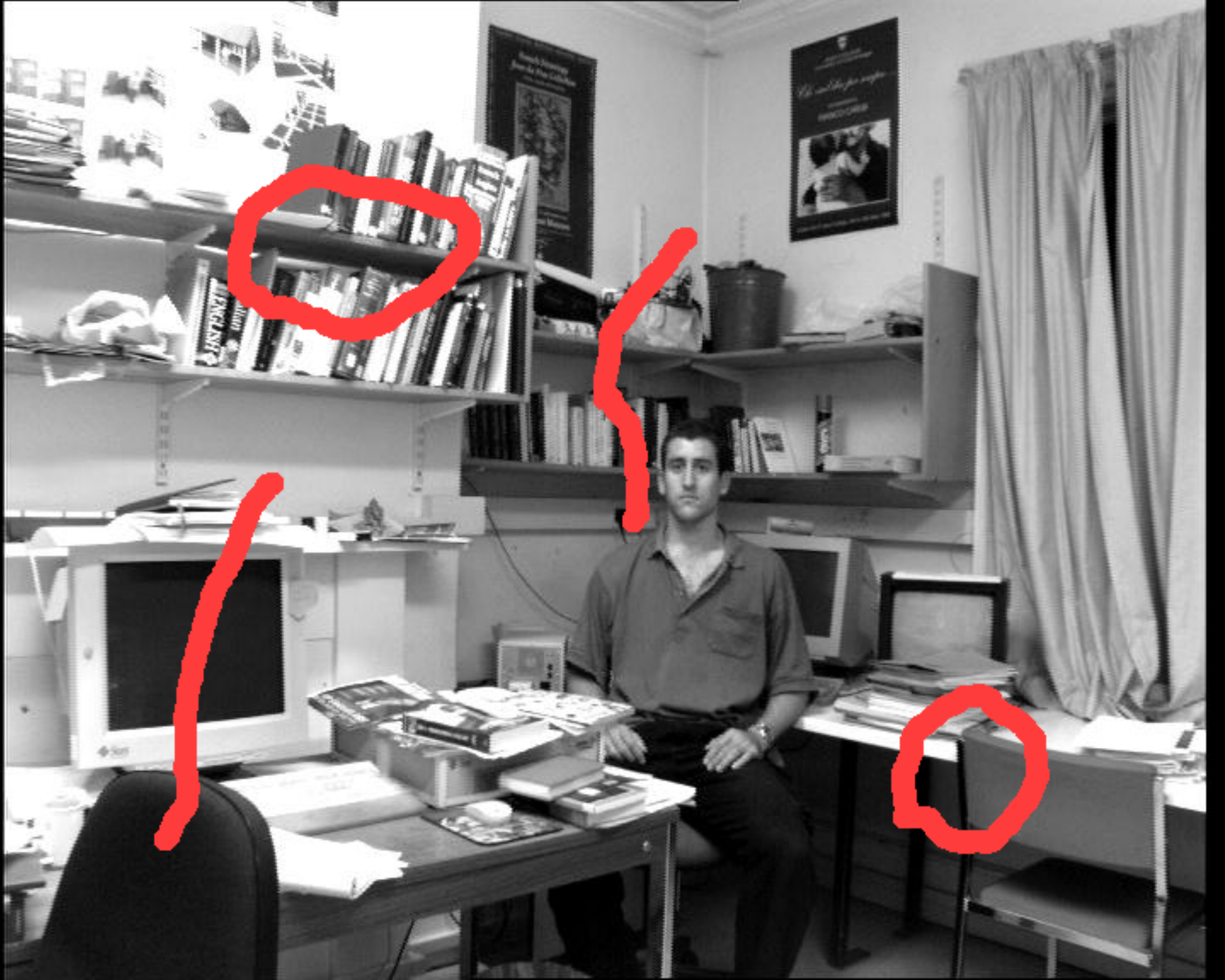} & 
\includegraphics[width=0.4\textwidth]{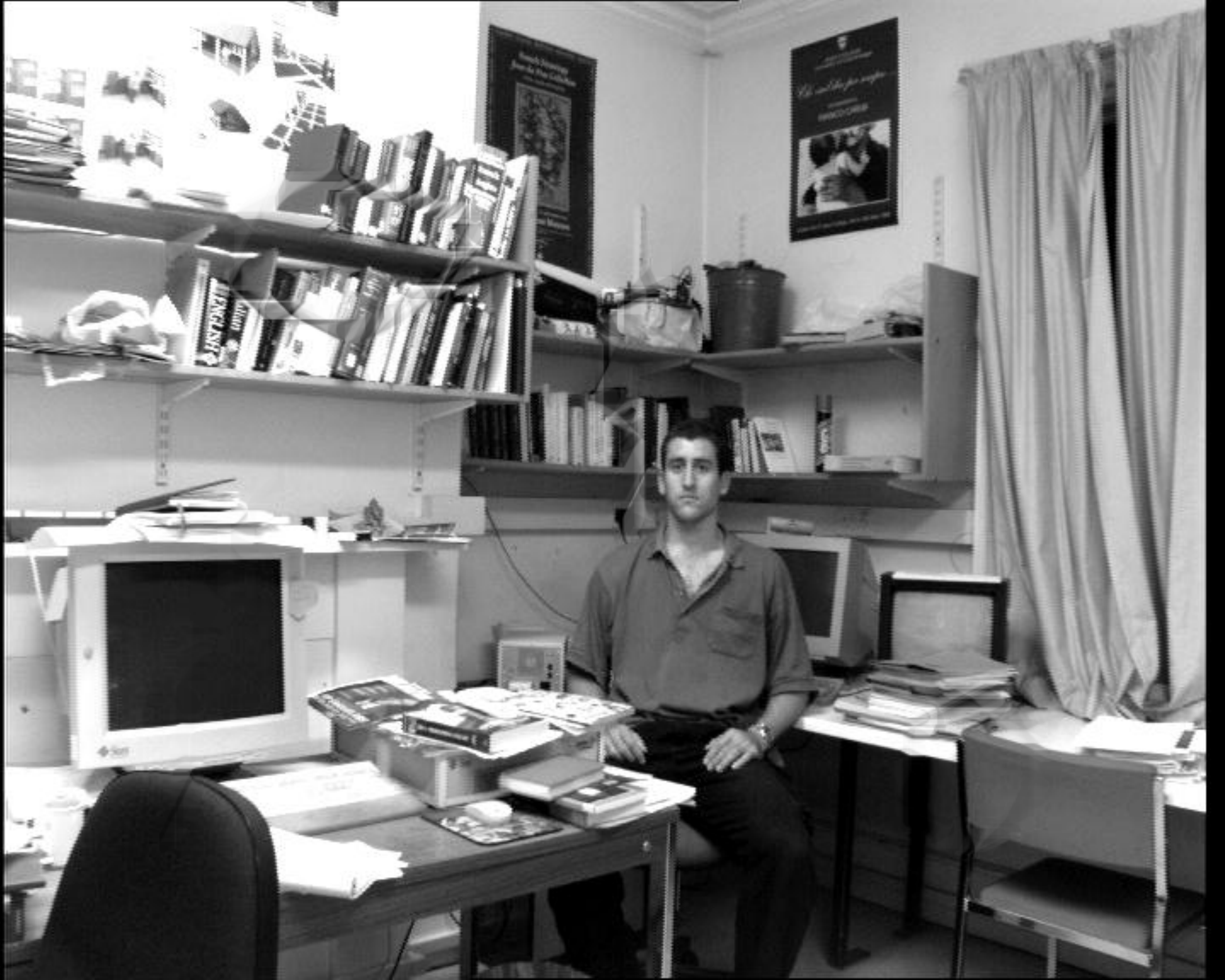}
\end{tabular}
\caption{Our method applied to structured inpainting. Images taken from \texttt{http://www.robots.ox.ac.uk/~vgg/data/data-various.html}.}
\label{fig:face_inpainting}
\end{center}
\end{figure*}

We now turn to the problem of inpainting, where we use interior point
solvers.  Figures \ref{fig:bridge_inpainting} and
\ref{fig:face_inpainting} show that our method is well-suited for
structured inpainting, and that length regularity generally does not
work well.

\begin{figure}
\begin{center}
\begin{tabular}{cc}
\includegraphics[width=0.46\linewidth]{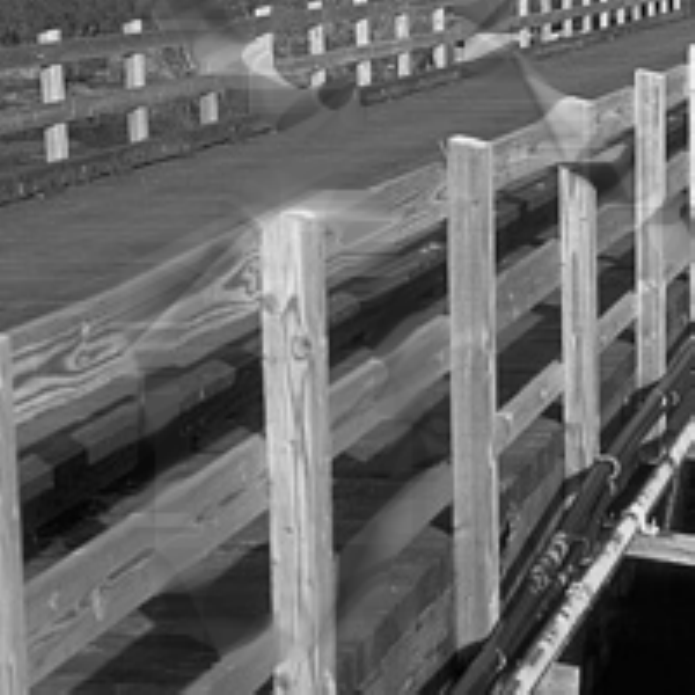} &
\includegraphics[width=0.46\linewidth]{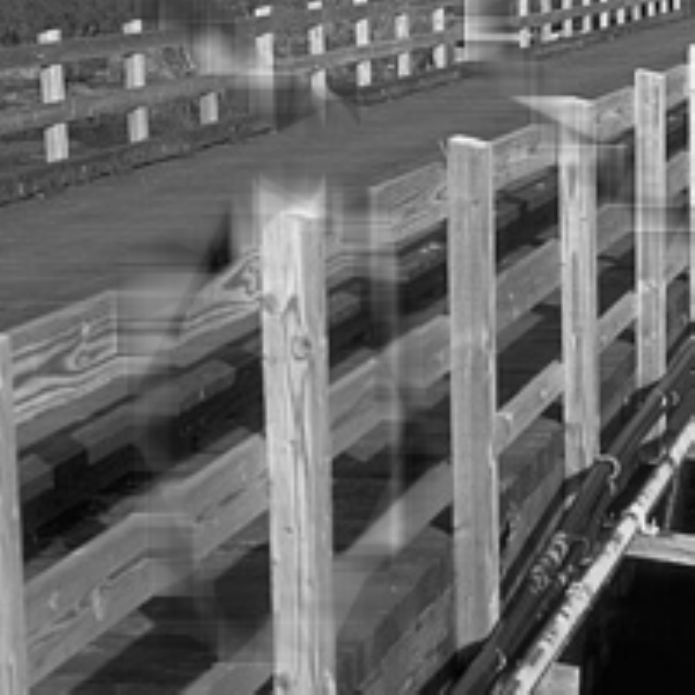} \\
\end{tabular}
\caption{Comparison of the proposed inpainting method (left column) and the one we proposed
in \cite{Schoenemann-Kahl-Cremers-09} (right column). The domain is as in Figure \ref{fig:bridge_inpainting}.}
\label{fig:inp_comparison}
\end{center}
\end{figure}

In this work we have improved upon our work
\cite{Schoenemann-Kahl-Cremers-09} by previously estimating the
direction of incoming level lines and giving a tighter constraint
system.  Figure \ref{fig:inp_comparison} shows that these changes
really improve the results.

\section{Conclusion}

We have presented new theory and methods for length- and
curvature-based regularization, both for image segmentation and
inpainting.  For curvature (in a region-based context) we are the
first to propose a global approach in the sense that it is independent
of initialization.

The results clearly demonstrate that curvature regularity outperforms
length-regularity in the presence of long and thin
objects. Experimentally we showed that for segmentation our strategy
of solving a linear programming relaxation is usually within $5\%$ of
the global optimum. In some cases it even finds the global optimum.

\section*{Acknowledgments}

We thank Petter Strandmark and Yubin Kuang for helpful
discussions.

Thomas Schoenemann and Fredrik Kahl were funded by the Swedish
Foundation for Strategic Research (SSF) through the programmes Future
Research Leaders and Wearable Visual Information Systems, and by the
European Research Council (GlobalVision grant no.\ 209480). Simon
Masnou would like to acknowledge funding by the French ANR
\emph{Freedom} project.  Daniel Cremers' work was financed through the
ERC starting grant ``ConvexVision''.

{\small
\bibliographystyle{plain}
\bibliography{cvrefs}
}
\end{document}